\documentclass[journal]{journal}
\ifCLASSINFOpdf
\else
\fi

\usepackage{amsmath,amssymb,amsfonts}
\usepackage{algorithmic}
\usepackage[linesnumbered,ruled]{algorithm2e}
\usepackage{graphicx}
\usepackage{subcaption} 
\usepackage{textcomp}
\usepackage{color}
\usepackage{mathtools}
\usepackage{bm}

\begin{document}
%
\title{Fast Approximate Bayesian Contextual Cold Start Learning (FAB-COST)}
%
%
%

\author{Jack R. McKenzie, Peter A. Appleby, Thomas House, Neil Walton
\thanks{Jack R. McKenzie, is a PhD student in Applied Mathematics at The University of Manchester, Manchester, UK. (Corresponding Author: jack.mckenzie@manchester.ac.uk)  }
\thanks{Neil Walton and Thomas House are both Readers in the School of Mathematics at The University of Manchester, Manchester, UK.}
\thanks{Peter A. Appleby is Head of Data Science at Auto Trader, Manchester, UK.  }}

\maketitle
\thispagestyle{empty}

%
\IEEEpeerreviewmaketitle

\begin{abstract}
Cold-start is a notoriously difficult problem which can occur in recommendation
systems, and arises when there is insufficient information to draw inferences for users or items. To address this challenge, a contextual
bandit algorithm -- the Fast Approximate Bayesian Contextual Cold
Start Learning algorithm (FAB-COST) -- is proposed, which is designed to provide improved
accuracy compared to the traditionally used Laplace approximation in the logistic contextual
bandit, while
controlling both algorithmic complexity and computational cost.  To this end,
FAB-COST uses a combination of two moment projection variational methods:
Expectation Propagation (EP), which performs well at the cold start, but
becomes slow as the amount of data increases; and Assumed Density Filtering
(ADF), which has slower growth of computational cost with data size but
requires more data to obtain an acceptable level of accuracy. By
switching from EP to ADF when the dataset becomes large, it is able to exploit
their complementary strengths. The empirical justification
for FAB-COST is presented, and systematically compared to other approaches on simulated
data.  In a benchmark against the Laplace approximation on real data consisting
of over $670,000$ impressions from autotrader.co.uk, FAB-COST demonstrates at one point
increase of over $16\%$ in user clicks. On the basis of these results,
it is argued that FAB-COST is likely to be an attractive approach to cold-start
recommendation systems in a variety of contexts.

\end{abstract}



\section{Introduction}

When making recommendations to website users, it is important to learn as
efficiently as possible which content is most appropriate to display, including
the `cold-start' case when there is little or no prior history of the user
and/or the content. Content recommendation systems and online advertising are
examples contexts, in which there is an intrinsic trade-off between
\emph{exploiting} current knowledge by e.g.\ displaying adverts believed most likely to be clicked on, and \emph{exploring} other content that might
have higher rewards by e.g.\ displaying adverts which there is currently little
information about.

Multi-armed bandits (MABs) are a class of algorithm which aim to balance the
exploration-exploitation dilemma present whenever an intelligent system must
make decisions in an uncertain environment \cite{chen2013combinatorial}. For a
recent and detailed overview on Multi-armed bandits see Lattimore \& Szepesvari
(2018) \cite{lattimore2018bandit}.  The most effective and conceptually simple
MAB algorithm is known as Thompson Sampling \cite{thompson1933likelihood},
which uses sampling from a posterior distribution obtained through Bayesian
inference. While Thompson's original (non-contextual) model for such inference
has the benefit of being analytically tractable, it has the drawback of each
action being assumed independent; when working in a large action space, it will
be much more efficient to share information among similar content, which is the
main motivation behind the contextual bandit.  Chapelle et al
\cite{chapelle2011empirical} propose such a contextual bandit based on Bayesian
logistic regression, which is not in general analytically tractable, leading to
the authors' use of the Laplace approximation.


Making use of the Bernstein-von Mises
and central limit theorems \cite{doob1949application}, the Laplace
approximation will have errors that are asymptotically $\mathrm{O}(T^{-1})$, where $T$ is number of impressions observed.
For extremely large datasets, these errors will therefore become negligible,
however they may be very large early on in the learning process, and as is shown, these errors may be large enough to have significant practical consequences
even after many thousands of observations.

The accuracy of the inference procedure is improved upon by using a combination of
Expectation Propagation (EP), which was contemporaneously devised by Minka
(2001) \cite{minka2001expectation} and Opper \& Winther (2000)
\cite{opper2000gaussian}, and Assumed density filtering (ADF) as presented by
Opper (1998) \cite{opper1998bayesian}.  These are both moment projection
variational inference methods, meaning that they work by iteratively projecting
the intractable posterior distribution onto a tractable one (usually belonging
to the exponential family) via minimisation of the forward Kullback-Leibler
divergence. ADF is a one-pass, online method, and observations are processed
one-by-one, updating the posterior distribution which is then approximated
before processing the next observation. EP -- which is an extension to ADF --
iteratively refines the approximation by making additional passes through the
dataset giving much better accuracy, but at the same time incurring a greater
computational cost.

Another class of methods considered are Markov chain Monte Carlo
(MCMC), which are capable of generating arbitrarily accurate representations of
the Bayesian posterior. These are useful to provide a `ground truth' for
comparison of approximate methods in a study such as this, however they are not
suitable for online use. This is because MCMC requires large amounts of
computational effort, and also typically involves algorithmic parameters that
currently cannot be tuned automatically, but rather need to be adjusted until
convergence can be diagnosed \cite{Brooks:2011}. 

The paper is structured as follows. Section \ref{sec:bandits} provides details
of the multi-armed bandit approach to recommender systems, and Section
\ref{sec:inference} provides details of the inferential systems used in
Bayesian online learning. In Section \ref{sec:fabcost} the
FAB-COST algorithm is introduced, which is systematically compared to Laplace, EP, ADF and
MCMC inference procedures, showing an attractive balance of accuracy and computational effort. It is then demonstrated that the increased accuracy of the posterior leads to better
performance when used in the logistic bandit setting and results in
more clicks generated when used in an online advertising scenario on real data
from autotrader.co.uk. Finally the details are concluded on in Section
\ref{sec:discuss}.

The code used to generate the results in this paper are available on GitHub at
https://github.com/JackMack21/FAB-COST.

\section{Bandit algorithms}

\label{sec:bandits}

\subsection{Notation and general setup}

Here and throughout, the shorthand $\mathcal{N}(\bm{\mu},\bm{\Sigma})$
represents the Gaussian distribution, where $\bm{\mu} \in \mathbb{R}^D$ is the
\emph{mean vector} of first moments and $\bm{\Sigma} \in \mathbb{R}^{D \times
D}$ is the \emph{covariance matrix} of centred second moments.
$\mathbb{E}_{p(x)}[\cdot]$ represents an expectation over the probability
distribution $p(x)$. 

Steps in the bandit algorithm are called \emph{iterations}, and these are indexed
by integers $i = 1, \ldots, T$. $T$ is the total
iterations over the learning process, and $\tau \leq T$ refers to a
current, but not necessarily final, iteration.  At each iteration, the bandit
algorithm --  also referred to as the learner, and which is
assumed to serve adverts for expositional simplicity -- selects an advert from
the set of eligible adverts $\mathcal{A}^i$, with cardinality
$|\mathcal{A}^i|=K$. This set is indexed by $j = 1, \ldots, K$. Each advert has $D$ features, examples of
which in the context of automobile sales being the colour, model and age of the
car advertised. As such, the options available can be represented as a matrix
$\mathbf{A}^i \in \mathbb{R}^{K \times D}$, with the $j$-th row,
$\mathbf{a}_j^{\top}$, having elements corresponding to the features of an
eligible advert.

%
%


The observed reward of the advert selected by the learner at iteration $i$ is
the binary outcome of a non-click/click, $y_i \in \{ 0, 1 \}$.
Each $y$ is treated as a random variable, the expected value for which at the
$i$-th iteration corresponds to the selected advert's click-through-rate (CTR).

$\mathbf{X} = [\mathbf{x_1}, \dots, \mathbf{x_T}]^\top $ is called the
\emph{design matrix} for the entire learning process. Each row
$\mathbf{x}_i^\top \in \mathbb{R}^D $ represents the features of the advert
displayed at the $i$-th iteration. For some algorithms the matrix $\mathbf{X}^\tau = [\mathbf{x_1}, \dots, \mathbf{x_\tau}]^\top$ will be used; 
this contains the information about the history of adverts chosen up to the current
iteration $\tau$. The history of observations up to iteration $\tau$ is
represented by the vector $\mathbf{y}_\tau \in \mathbb{R}^\tau$.
 
$\bm{\theta} \in \Theta$ is the vector of parameters that is to be learnt.
In the non-contextual case, each advert has a \emph{local} parameter $\theta_j$
that is learnt and so $\Theta = \mathbb{R}^K$. In the contextual case $\Theta =
\mathbb{R}^D$ and $\bm{\theta}$ is a \emph{global} parameter vector which
shares information between adverts. 

\subsection{Thompson Sampling}

Thompson sampling \cite{thompson1933likelihood} is a Bayesian approach to MABs. Its use requires the quantification of belief about the CTR for each advert via a posterior probability
distribution. The adverts shown and the binary reward of click/no-click
observed up to the current time $\tau$ are denoted via the vectors
$\mathbf{a}_\tau = [a_1, \dots, a_\tau]^\top$ and $\mathbf{y}_\tau =
[y_1,\dots, y_\tau]^\top$ respectively. As data arrives, which is comprised of
the tuple $(a_i,y_i)$, the latent parameter
$\bm{\theta}$ can be learnt via Bayes' rule
\begin{equation*}
p(\bm{\theta}|\mathbf{y}_{\tau},\mathbf{a}_\tau) \propto
\prod_{i=1}^\tau p(y_i|a_i,\bm{\theta}) p(\bm{\theta}).
\end{equation*}
Once a posterior distribution has been calculated for each advert, an advert should be chosen which maximises the expected CTR, where the expected CTR is
calculated as
\begin{equation}
\begin{multlined}
\mathbb{E}[y_{\tau+1}|a_{\tau+1},\mathbf{a}_\tau,\mathbf{y}_\tau]
= \\ \int
\mathbb{E}[y_{\tau+1}|\mathbf{y}_\tau,a_{\tau+1},\mathbf{a}_\tau,\bm{\theta}]p(\bm{\theta}|\mathbf{y}_\tau,a_{\tau+1},\mathbf{a}_\tau)
{\rm d}\bm{\theta}.
\label{eq:expected_reward2}  
\end{multlined}
\end{equation}
In Thompson sampling it is not necessary to evaluate this integral, but rather a sample is generated from each advert's CTR posterior, and the advert
which corresponds to the maximum sample generated is displayed. Although the integral
\eqref{eq:expected_reward2} could be evaluated explicitly, displaying content
that simply maximises it would involve constantly exploiting existing knowledge
and would never explore to learn. The sampling-based approach, however, allows
exploration and exploitation to be traded off 

A well studied example of Thompson Sampling is the non-contextual
Beta-Bernoulli bandit. In this simple case, each advert is assumed to have an
independent Bernoulli likelihood. This has a conjugate prior, the Beta
distribution, meaning that Bayesian inference can be performed analytically to
give a closed form solution.  When working in a very large action space (i.e.\
with many adverts) it is, however, much more efficient to share information
among similar adverts rather than assuming independence, which is the main
motivation behind the contextual bandit. 

\subsection{The Contextual Bandit}

Instead of learning an independent posterior distribution for each advert, the
contextual bandit instead learns a \textit{global} posterior
$p(\bm{\theta}|\mathbf{X})$ where $\bm{\theta} \in \mathbb{R}^D$, with $D$
representing the number of covariates. 

As discussed above, at the current iteration $\tau$, the learner is presented
with the features of the adverts available, stored in the matrix
$\mathbf{A}^\tau$.  The learner then chooses an advert from $\mathcal{A}^\tau$
which is expected to have the highest CTR when combining the context received
with the sample generated from the posterior 
\begin{equation}
\label{eq:contextual_TS1}
a_\tau = \mathrm{argmax} (\mathbf{A}^i \bm{\theta}),
\end{equation}
i.e.\ if the $j$-th row of $\mathbf{A}^i \bm{\theta}$ as in
\eqref{eq:contextual_TS1} is the maximum, advert $a_j$ is shown. After observing
the reward $y_i$, the context of the advert chosen $a_j$ is then added to the
history of chosen adverts (corresponding to the $i^{\text{th}}$ row of
$\mathbf{X}$), and the posterior is updated either in batch using
$\mathbf{X}^\tau$ or online using the chosen adverts covariates which is now
denoted $\mathbf{x}_\tau$.

The global parameter vector $\bm{\theta}$ therefore acts as a projection of the
features onto the real numbers. Since the outcome is a binary reward
$y_i \in \{0, 1\}$, an approach is needed that relates a continuous projection to
such an outcome. Bayesian logistic regression fulfils this requirement, which is the next point of discussion.  

\subsection{Bayesian Logistic regression}

Logistic regression has proven to be a very popular and effective two class
`soft' classification method \cite{Hosmer:2004,Hastie:2009}. The goal of
logistic regression is to find the best fitting model to describe the
relationship thus far observed between the binary response variables
$\mathbf{y} \in \mathbb{R}^{\tau}$, and the design matrix $\mathbf{X} \in
\mathbb{R}^{\tau \times D}$.  Although in statistics literature it is referred
to as logistic regression, by virtue of modelling a binary outcome it is also a
method for classification \cite{bishop2006pattern}.  While there has recently
been much interest in learning techniques based on e.g.\ neural networks, as
Dacrema et al.\ \cite{dacrema2019we} point out, interpretability and
reproducibility are two very important issues that need to be addressed, and
are important advantage of using well-understood statistical techniques. 

Logistic regression is a likelihood-based method in which the parameter vector
$\bm{\theta}$ describes the probabilistic relationship between the input vector
$\mathbf{x}_i$, and a binary response $y_i \in \{0 ,1\}$. Due to the response
being binary, a Bernoulli model is used for probabilities:
\begin{equation}\label{eq:log_reg_prob}
\mathrm{Pr}(y_i|\mathbf{x}_i,\bm{\theta})
= \pi(\mathbf{x}_i, \bm{\theta})^{y_i} (1-\pi(\mathbf{x}_i,
	\bm{\theta}))^{(1-y_i)},
\end{equation}
where $\pi(\mathbf{x}_i, \bm{\theta}) = \mathbb{E}[y_i|\mathbf{x}_i,
\bm{\theta}] = p(y_i=1|\mathbf{x}_i, \bm{\theta})$. As $\pi(\mathbf{x}_i, \bm{\theta})$ is
the probability of observing a positive outcome, the inner product
$\bm{\theta}^\top\mathbf{x}_i $ is mapped from the real line to the interval $[0, 1]$
via the (sigmoidal) \emph{logistic} function, $\sigma(x) =
1/(1+\mathrm{exp}(-x))$,
\begin{equation}
\pi(\mathbf{x}_i, \bm{\theta}) = 
\sigma(\bm{\theta}^\top\mathbf{x}_i) = \frac{1}{1+
{\rm exp}(-\bm{\theta}^\top\mathbf{x}_i)}. \label{eq:logistic_function_def}
\end{equation}
Assuming conditional independence at each iteration, the
likelihood function for the current iteration is obtained as
\begin{equation}\label{eq:freq_log_reg_lilihood}
p(\mathbf{y}_\tau| \mathbf{X}^\tau, \bm{\theta}) =
\prod_{i=1}^{\tau}\mathrm{Pr}(y_i|\mathbf{x}_i,\bm{\theta}),
\end{equation} 
where the historical observations are represented as $\mathbf{y}_\tau = [y_1,
\dots , y_\tau]^\top $ and $\mathbf{X}^\tau = [ \mathbf{x}_1 , \dots ,
\mathbf{x}_\tau]^\top$. Equations \eqref{eq:log_reg_prob},
\eqref{eq:logistic_function_def} and \eqref{eq:freq_log_reg_lilihood} between
them define logistic regression.

To select adverts via Thompson sampling, $\bm{\theta}$ must be estimated,
including an appropriate quantification of uncertainty, which motivates the use
of a Bayesian treatment of logistic regression. Here, it is assumed that there is a
distribution over $\bm\theta$ that is sequentially learnt as data arrives via
Bayes' rule:
\begin{equation}
p(\bm{\theta}|\mathbf{y}_\tau,\mathbf{X}^\tau) \propto 
p(\mathbf{y}_\tau| \mathbf{X}^\tau, \bm{\theta})p(\bm{\theta}),
\label{eq:bayes}
\end{equation}
where; $p(\bm{\theta})$ is a probability density function representing the
\emph{prior} belief about the parameters,
$p(\bm{\theta}|\mathbf{y}_\tau,\mathbf{X}^\tau)$ is a probability density
function representing the \emph{posterior} beliefs about the parameters, and
all other quantities are as defined above.  Since $\bm{\theta}$ is passed
through a non-linear mapping, inference is not straightforward; more
explicitly, the logistic likelihood function does not permit a conjugate prior.
This leads onto the next section which explains methods for dealing with such situations.

\section{Inferential methodology}

\label{sec:inference}

Suppose that one is trying to solve the equation \eqref{eq:bayes} for the
posterior distribution -- dependence on
$\mathbf{y}_\tau,\mathbf{X}^\tau$ is suppressed the exact distribution is denoted
$p(\bm{\theta})$. When there is a conjugate prior, the constant of
proportionality in \eqref{eq:bayes} can be calculated exactly, but otherwise it
must be approximated. 

Markov chain Monte Carlo (MCMC) methods are very popular in Bayesian
statistics; and provided there is sufficient computational resources, they
guarantee an arbitrarily accurate approximation to the posterior distribution.
The need for large computational resources can, however, be problematic,
especially when working at scale. An overview of MCMC is provided by Brooks
et al.\ \cite{Brooks:2011}.

Other approaches that will be used and detail below make a Gaussian approximation,
which is justified via the Bernstein Von Mises Theorem. This states that the
posterior converges to a Gaussian asymptotically \cite{dasgupta2008asymptotic},
however different methods will achieve different accuracies at a given amount
of data.

\subsection{The Laplace approximation}

The Laplace approximation is a very popular inference method in Bayesian
statistics. Its popularity is due to its simplicity: given a target
distribution $p(\bm\theta) = \exp(-\xi(\bm\theta))$, which in this case will be
the posterior distribution, a tractable Gaussian approximation
$q(\bm\theta)$ is made, centred at the mode of the original target density with
variance equal to the curvature of the negative log-target. Explicitly,
let
\begin{equation*}
\bm\mu^* = \underset{\bm\theta}{\text{argmin}} \ \xi(\bm\theta), \qquad
\bm\Lambda^* = \left.
\frac{\partial^2 \xi(\bm\theta)}{\partial \bm\theta^2 }
\right|_{\bm\theta = \bm\mu^*},	
\end{equation*}
and then the two moments of the Gaussian approximation are matched to the
above:
\begin{equation*}
p(\bm\theta) \approx
q(\bm\theta) = \mathcal{N}(\bm\mu^*, {\bm\Lambda^*}^{-1}).
\end{equation*}
The Laplace approximation is used as an inference procedure in the logistic
contextual bandit described by Chapelle et al \cite{chapelle2011empirical} and
very commonly when making an approximation in Bayesian GLMs. Asymptotically,
the errors are expected to be $O(T^{-1})$ \cite{murray2012asymptotic}.

\subsection{Variational Inference}

Variational approximations turn what was an inference problem into one of
optimisation. They work by minimising a distance measure $\mathcal{D}$ between
the target distribution $p(\bm\theta)$ and an approximation $q(\bm\theta) \in
\mathcal{Z}$, where $\mathcal{Z}$ is the family of densities the approximation is to be restricted to, so that
\begin{equation}\label{eq:VI_optimisation}
q^*(\bm\theta) = \underset{q \in \mathcal{Z}}{\text{argmin}} \
	\mathcal{D}(q(\bm\theta),p(\bm\theta|\mathbf{x})).
\end{equation}
The more generality there is in $\mathcal{Z}$, the more complex the
optimisation procedure becomes, and in this application, the
approximating family is restricted to the set of Gaussians.

The distance measure used is typically the Kullback-Leibler divergence
\cite{kullback1951information}, also known as the relative entropy, which is
non-symmetric.  Here, the \emph{forward} KL-divergence, defined as
\begin{align}\label{eq:for_KL2}
 	\text{KL}(p(\bm\theta|\mathbf{x})||q(\bm\theta)) & = \int p(\bm\theta|\mathbf{x}) \log \left\{ \frac{p(\bm\theta|\mathbf{x})}{q(\bm\theta)}\right\} d\bm\theta\\	
 	& = \mathbb{E}_{p(\bm\theta|\mathbf{x})}\left[ \log\left( p(\bm\theta|\mathbf{x}) \right) - \log\left( q(\bm\theta) \right) \right]\nonumber
 \end{align} is used as the distance measure, which
when used in the objective in (\ref{eq:VI_optimisation}), gives a solution known
as the \emph{moment projection}.

Note that the \emph{reverse} KL-divergence is obtained by swapping $p$ and $q$
in \eqref{eq:for_KL2} and when used as the objective in
(\ref{eq:VI_optimisation}), the solution is known as the \emph{information
projection}. This form is not necessarily convex in $\bm\theta$ and can
therefore yield different solutions depending on how the optimisation procedure
is initialised. While the information projection is most commonly used in
variational inference due to its simplicity, there are problems with
reliability and accuracy, and no advantage was found by using this
approach -- see Bishop (2006) \cite{bishop2006pattern} and Blei et al.\ (2017)
\cite{blei2017variational} for further discussion. 

Unlike the information projection, the objective (\ref{eq:for_KL2}) is convex
in $\bm\theta$, and will give a unique solution when minimised. This minimum
corresponds to an approximation centred at the mean of the target which is why
it is known as \textit{mean seeking}.  The moment projection plays an important
role in asymptotic theory and as explained by
\cite{bernardo1987approximations}, its minimisation has the desirable effect of
minimising the expected loss. 
 
Although it is the `correct' KL-divergence measure to use, the forward version
has the drawback that it is much harder to compute; it is intractable as it
requires the calculation of the expectation over the target $p(\bm{\theta}|\mathbf{x})$.
EP overcomes the intractability of the target by forming what is known as a
tilted distribution; this will be discussed in the next section. 

\subsection{Exponential families}\label{sec:exp_fam}

A distribution belongs to the set of exponential families if its density can be
written as
\begin{align*}
	q(\bm\theta|\bm{\lambda}) & = \exp\left( \bm{\lambda}^T \phi(\bm\theta) - \Phi(\bm{\lambda}) \right), \hspace{1.5cm} \bm{\lambda} \in \bm{\Theta}\\
	\Phi(\bm{\lambda}) & = \log \int \exp \left( \bm{\lambda}^T \phi(\bm\theta) \right) {\rm d} \bm\theta	, 
\end{align*}
where $\phi(\bm\theta)$ is known as the sufficient statistics, $\bm\lambda$
the natural parameters and $\Phi(\bm{\lambda})$ is the log-partition function
which ensures normalisation. 

Assuming that the representation of the exponential family is minimal (i.e.
there are no dependencies between the components of $\phi(\bm\theta)$ and
$\bm\lambda$), then the following properties hold:

\subsubsection{Product of Exponentials}\label{sec:exp_properties_prod}

A product of densities belonging to the set exponential families is also a
member of the exponential families:
\begin{equation*}
\prod_{i=1}^T q(\bm\theta|\bm\lambda_i) = q\left( \bm\theta
\Big| \sum_{i=1}^T \bm\lambda_i \right) \exp \left( \Phi \left(
\sum_{i=1}^T \bm\lambda_i \right) - \sum_{i=1}^T
\Phi(\bm\lambda_i) \right),
\end{equation*}
given that $\sum_{i=1}^T \bm\lambda_i \in \bm\Theta$.

\subsubsection{Moments}\label{sec:exp_properties_moments}

Moments can be found by differentiating the log-partition function with respect to its natural parameters:
\begin{equation*}
	\mathbb{E}_{q(\bm\theta)}[\phi(\bm\theta)] = \nabla_{\bm\lambda} \Phi(\bm\lambda),\quad
	\text{Var}_{q(\bm\theta)}[\phi(\bm\theta)] = \nabla^2_{\bm\lambda} \Phi(\bm\lambda).
\end{equation*}

\subsubsection{Bijective mapping}\label{sec:exp_properties_mapping}

There is a bijective mapping from the natural parameters $\lambda(\bm\eta)$ and
the moment parameters $\bm\eta(\bm\lambda)$.  The log-partition function
$\Phi(\bm\lambda)$ is strictly convex and has a Legendre dual 
\begin{equation*}
	\Psi(\bm\eta) = \mathbb{E}_{\eta(\bm\lambda)} [\log p (\lambda(\bm\eta)|\bm\theta)]
\end{equation*}
Conversion between the natural and moments parameters is done via:
\begin{equation*}
\bm\eta(\bm\lambda) = \nabla_{\bm\lambda} \Phi(\bm\lambda), \qquad \theta(\bm\eta) = \nabla_{\bm\eta}
	\Psi(\bm\eta)
\end{equation*}

These properties are very useful for message passing algorithms which both EP
and ADF are a subset of -- see Pearl (1986) \cite{pearl1986fusion}. Seeger
(2007) \cite{seeger2005expectation} gives an in-depth discussion of the
properties of exponential families.

\section{FAB-COST}

\label{sec:fabcost}

In this section a description of the FAB-COST algorithm is given, along with a demonstration of its performance on real automotive website data (AT) taken from the autotrader.co.uk website. 

\subsection{The Logistic Contextual Bandit}

Having introduced multi-armed bandits, Thompson sampling for these via Bayesian
logistic regression, and multivariate normal approximations to the posterior in
such regressions, the overall structure of
the FAB-COST approach, as in Algorithm \ref{algorithm:posterior_sampling}, can now be provided.  It is worth
mentioning that the pseudocode for the logistic regression bandit provided is
the same as the linear bandit introduced by Russo et al (2014)
\cite{russo2014learning} and Agrawal et al (2013) \cite{agrawal2013thompson},
however the linear case is conjugate and can be solved exactly, meaning that it
does not require the work to update moments accurately that have been carried
out.

\begin{algorithm}[h!]
    \SetKwInOut{Input}{Input}
    \SetKwInOut{Output}{Output}
    \For{$ i = 1 \dots T$}
    {
    1. Generate a sample from the approximated posterior:\\ \hspace{1cm}$\tilde{\bm\theta}_i \sim \mathcal{N}(\bm\mu_{i-1},\bm\Sigma_{i-1})$\\
    2. Select an advert: \\
    \hspace{1cm}$a_i = \underset{\mathrm{ j \in \mathcal{A}}}{\text{argmax}} (\mathbf{A}^i \tilde{\bm{\theta}}_i)$\\
    3. Update moments:\\
    \hspace{1cm}$\bm\mu_{i} = \mathbb{E}[\bm\theta|\mathbf{X}_i, \mathbf{y}_i] $\\
    \hspace{1cm}$\bm\Sigma_{i} = \mathbb{E}[(\bm\theta -\bm\mu_i)(\bm\theta -\bm\mu_i)^\top|\mathbf{X}_i, \mathbf{y}_i]  $
    }
    \caption{Logistic Regression Thompson Sampling}\label{algorithm:posterior_sampling}
\end{algorithm}		

Describing this algorithm in long form, it is initialised with a prior belief on
$\bm{\theta}$. At each iteration, the learner is presented with an available
set of adverts and random sample is generated from the posterior distribution
(or prior at the first iteration), which corresponds to step 1 in Algorithm
\ref{algorithm:posterior_sampling}. An advert is then selected from the set via
Thompson Sampling by choosing $a_j$ which maximises the linear combination of
the sample generated $\tilde{\bm{\theta}}_i$ and the eligible adverts
covariates $\mathbf{A}^i$; this corresponds to step 2 in Algorithm
\ref{algorithm:posterior_sampling} -- note that the monotonicity of the
logistic function means that the sampled CTR is maximised when the
linear combination is maximised.  This is followed by observing a binary reward of a user
clicking or not clicking on the chosen advert, at which point the posterior
$p(\bm{\theta})$ is updated (i.e.\ within the Gaussian approximation, the first
and second moments are updated). This corresponds to step 3 in Algorithm
\ref{algorithm:posterior_sampling}.

\subsection{Expectation Propagation}


Expectation Propagation (EP) is an iterative approach to minimising the forward
KL-divergence between the posterior that is to be approximated, and the
Gaussian approximation.  It was first generalised by Minka (2001)
\cite{minka2001expectation} and Opper \& Winther (2000)
\cite{opper2000gaussian}, but has roots further back in Statistical Physics
\cite{mezard1985random}.  EP belongs to a group of message passing algorithms
and works by essentially propagating the moments of an exponential family -
which in the Gaussian case are the mean and variance - between the factors of
the posterior. Finding the moments of the target is obviously problematic as
the target is intractable - if it weren't then a closed form
solution could be found analytically. EP's solution to this is to form what is known as the
\textit{tilted distribution} $t_i(\bm{\theta})$ (whose
moments are much easier to find) and iteratively project the moments from this,
onto the tractable approximation $q(\bm\theta)$. 

First, it is assumed that the true posterior factorises into a product of $T$ factor
terms or \textit{sites}:
\begin{equation*}\label{eq:target}
	p(\bm\theta|\mathbf{x}) \propto \prod_{i=1}^T p_i(\bm\theta) .
\end{equation*} 
EP approximates each of these true sites by a Gaussian distribution
$q_i(\bm\theta)$, which in natural parameters is expressed as
\begin{equation*}
p_i(\bm\theta) \approx q_i(\bm\theta|\bm\lambda_i) \propto \exp \left\{ \mathbf{h}_i
	\mathbf{x}_i - \bm{\Lambda}_i \frac{\mathbf{x}_i^2}{2}  \right\}.
\end{equation*}
Then due to property 1 in of exponential families in \S{}\ref{sec:exp_fam}, the
global approximation is expressed as
\begin{equation*}\label{eq:global_approx}
	q(\bm\theta|\bm\lambda) = \prod_{i=1}^T q_i(\bm\theta|\bm\lambda_i),
\end{equation*}
and the natural parameters of our global approximation can be calculated as the
product of the natural parameters of each site approximation $\mathbf{h} =
\sum_{i=1}^T \mathbf{h}_i$, $\bm{\Lambda} = \sum_{i=1}^T \bm{\Lambda}_i$. It is
this ability to simply add and subtract natural parameters of the sites that
motivates the use of an exponential family approximation. 

The EP algorithm sweeps through the data set, with steps described below.

\subsubsection{The tilted distribution}

At each iteration of the algorithm, the current \textit{global
approximation} $q(\bm\theta|\bm\lambda) = \prod_{i=1}^T q_i(\bm\theta|\bm\lambda_i)$ is augmented by replacing one
of the sites with a true site $p_i(\bm\theta)$.  This can be thought of in two
steps: firstly the \textit{cavity distribution} is defined as
\begin{equation*}
q^{\setminus i}(\bm\theta|\bm\lambda^{\setminus i}) = \prod_{j \neq i} q_j(\bm\theta|\bm\lambda_j),
\end{equation*}
which is the global approximation with a site removed, then the
\textit{tilted distribution} is defined as
\begin{equation*}\label{eq:tilted}
t_i(\bm\theta)  \propto p_i(\bm\theta) \prod_{j \neq i} q_j(\bm\theta|\bm\lambda_j),
\end{equation*}
which is the cavity with its `hole' filled in with a true site.

\subsubsection{The moment projection}

EP proceeds to iteratively project the tilted onto the global approximation 
\begin{equation}\label{eq:EP_objective}
q^*(\bm\theta) = \underset{\mathrm{ q \in \mathcal{Z}}}{\text{argmin}} \
	\text{KL}(t_i(\bm\theta)||q(\bm\theta|\bm\lambda)).
\end{equation}
Equivalently, the first two moments of the tilted can be computed $\mathbb{E}_{t_i}[\phi(\bm\theta)] = [\bm\mu,\bm\Sigma]^\top$ and the
moments of the global approximation are equated to these: $q^*(\bm\theta)\sim
\mathcal{N}(\bm\mu,\bm\Sigma).$ 

The moments of the local site approximation $q_i(\bm\theta_i)$ must then be updated which is generally done by division
\begin{equation*}\label{eq:site_update}
q^*_i(\bm\theta|\bm\lambda_i) =
\frac{q^*(\bm\theta|\bm\lambda)}{q^{\setminus i}(\bm\theta|\bm\lambda^{\setminus i})}.
\end{equation*}
Using property 1 in of exponential families in \S{}\ref{sec:exp_fam}, this
results in a simple subtraction of the natural parameters. 

\subsubsection{Mapping from natural to moment parameters}

As the objective in (\ref{eq:EP_objective}) is found via moment matching, yet
the Gaussian approximation is parameterised by its natural parameters, the two different parameterisations but be alternated between at each iteration.
Due to property 3 of exponential families in \S{}\ref{sec:exp_fam}, there is a
bijective mapping between the two. In the Gaussian case these mappings are:
\begin{equation}\label{eq:moment_mapping}
\boldsymbol{\Sigma}  = \boldsymbol{\Lambda}^{-1}, \qquad
\boldsymbol{\mu}  = \boldsymbol{\Lambda}^{-1} \mathbf{h},
\end{equation}
where the moment parameters -- the mean and the variance -- $\bm\mu$ and
$\bm\Sigma$ respectively, and the natural parameters -- the shift and the
precision -- are given by $\mathbf{h}$ and $\bm\Lambda$ respectively.

\subsubsection{Overall structure}
 
As shown in Algorithm \ref{alg:EP}, EP makes multiple sweeps through the
dataset, iteratively forming the tilted distribution, matching the moments of
the global approximation, and finally updating the site parameters. This is
done until convergence which has been shown to usually be after 3-4 sweeps
(which coincides with the experiments carried out in this work) although a stopping rule described by
Seeger (2008) \cite{seeger2008bayesian} can also be used. 

 \begin{algorithm}
 \While{not converged}
 {
 \For{$ i = 1 \dots T$}
 {1. Form the tilted distribution:\\ \hspace{1cm} $t_{i}(\bm\theta) = \frac{1}{\tilde{Z}}p_i(\bm\theta) q^{\setminus i}(\bm\theta|\bm\lambda) $.\\
 2. Minimise the forward KL-divergence between the tilted distribution and the global approximation:\\ \hspace{1cm} $q^{*}(\bm\theta|\bm\lambda) = \underset{\mathrm{ q \in \mathcal{Z}}}{\text{argmin}} \ \text{KL}(t_i(\bm\theta)||q(\bm\theta|\bm\lambda)).$ \\
 3. Update the approximating site:\\ \hspace{1cm} $q^*_i(\bm\theta|\bm\lambda_i) = \frac{q^*(\bm\theta|\bm\lambda)}{q^{\setminus i}(\bm\theta|\bm\lambda^{\setminus i})}$.
 } }
\caption{The Expectation Propagation algorithm}\label{alg:EP}
\end{algorithm} 

\subsubsection{Error analysis}

While EP has shown tremendous empirical success in many applications --
particularly Bayesian General Linear Models and Gaussian Process regression
\cite{gelman2014expectation} -- the theoretical understanding of it is less
comprehensive than for other approaches. There has, however, been important
progress due to Dehaene and Barthelm\'{e} who show that EP behaves like
iterations of Newton’s algorithm for finding the mode of a function
\cite{dehaene2018expectation}. Under what they describe in later work as
`unrealistic assumptions on the model', Dehaene and Barthelm\'{e} also showed
that EP can converge at a rate of $O(T^{-2})$ \cite{dehaene2015bounding}. The
experiments carried out in this work suggest that while such a rate is indeed likely to be optimistic in
more realistic settings, EP is often significantly more accurate than the
Laplace approximation.

\subsection{Assumed density filtering}

\subsubsection{Algorithm}

EP is a batch method, requiring multiple sweeps through a complete dataset,
potentially limiting its usefulness online. Assumed Density filtering (ADF) is
a sequential inference method and can be used to work online. As with EP, ADF
iteratively minimises the forward KL-divergence between the tilted distribution
and the approximation, the difference being how this tilted distribution is
formed. 

In the setting of data arriving sequentially, the posterior distribution at
data point $\tau$ is given as 
\begin{equation*}
	p(\bm\theta|\mathbf{x}) = \frac{\prod_{i=1}^\tau
	p(\mathbf{x}_i|\bm\theta)p(\bm\theta)}{\int \prod_{i=1}^\tau
	p(\mathbf{x}_i|\bm\vartheta)p(\bm\vartheta) {\rm d}\bm\vartheta}.
\end{equation*}
ADF takes $q(\bm\theta)$ as the prior on $\bm\theta$ and iterates through the
data, incorporating each point into the approximate posterior. The conditional
distribution of $\bm\theta$ given the first $\tau$ data points can be expressed
as
\begin{equation}\label{eq:condtional_ADF}
p(\bm\theta|x_{1:\tau}) =
\frac{p(\mathbf{x}_\tau|\bm\theta)p(\bm\theta|\mathbf{x}_{1:\tau-1})}{\int
p(\mathbf{x}_\tau|\bm\vartheta)p(\bm\vartheta|\mathbf{x}_{1:\tau-1})
	{\rm d}\bm\vartheta}.
\end{equation}
Then assuming that at the previous iteration the approximation
$q^{(\tau-1)}(\bm\theta)$ is made to the true posterior
$p(\bm\theta|\mathbf{x}_{1:\tau-1})$, \eqref{eq:condtional_ADF} can be rewritten
to give a new \textit{tilted distribution}
\begin{equation*}\label{eq:hybrid_ADF}
	t_\tau(\bm\theta|\mathbf{x}_{\tau}) =
	\frac{p(\mathbf{x}_\tau|\bm\theta)q^{(\tau-1)}(\bm\theta)}{\int
	p(\mathbf{x}_\tau|\bm\vartheta)q^{(\tau-1)}(\bm\vartheta) 
	{\rm d}\bm\vartheta}.
\end{equation*}
Due to there being neither a site update nor a cavity there is no need to map
between the natural and moment parameters and therefore no need to perform the
expensive matrix inversion seen in equation \eqref{eq:moment_mapping}.

 

\subsubsection{Error analysis}

\begin{figure}[h!]
     \begin{subfigure}[h!]{0.23\textwidth}
        \includegraphics[width=\linewidth]{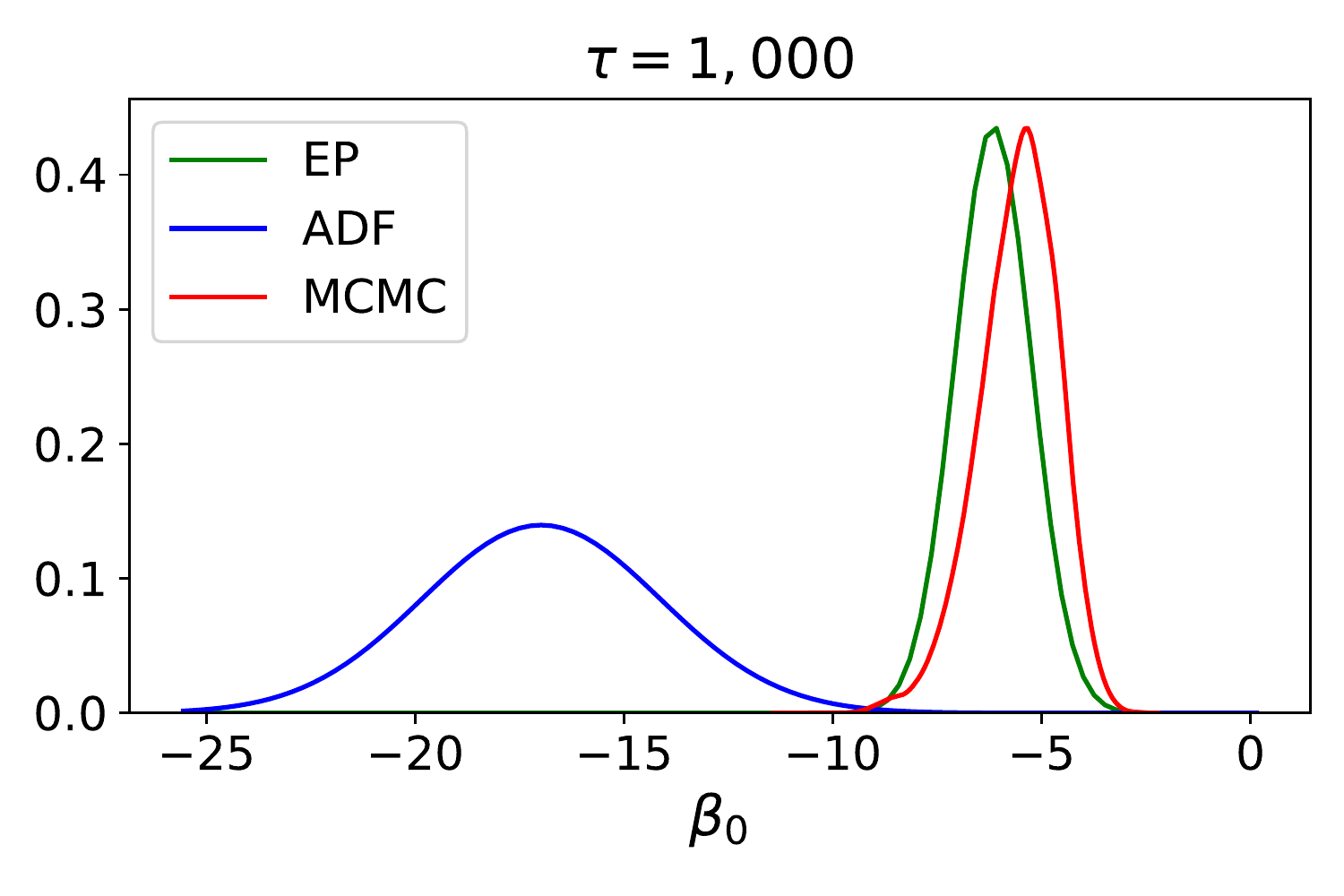}
        \label{fig:b0_1000}
    \end{subfigure}
    \begin{subfigure}[h!]{0.23\textwidth}
        \includegraphics[width=\linewidth]{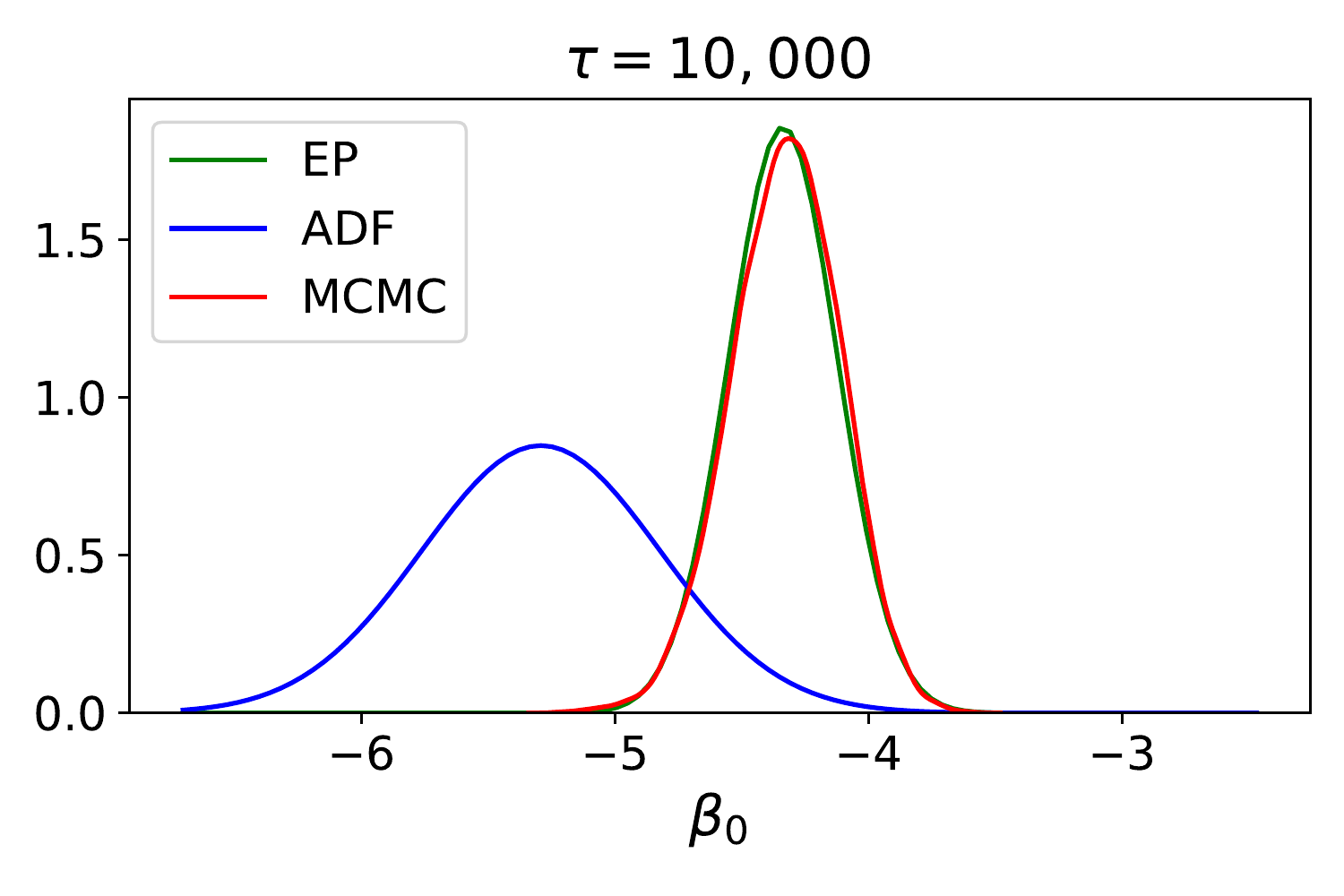}
        \label{fig:b0_10000}
    \end{subfigure}
    \begin{subfigure}[h!]{0.23\textwidth}
        \includegraphics[width=\linewidth]{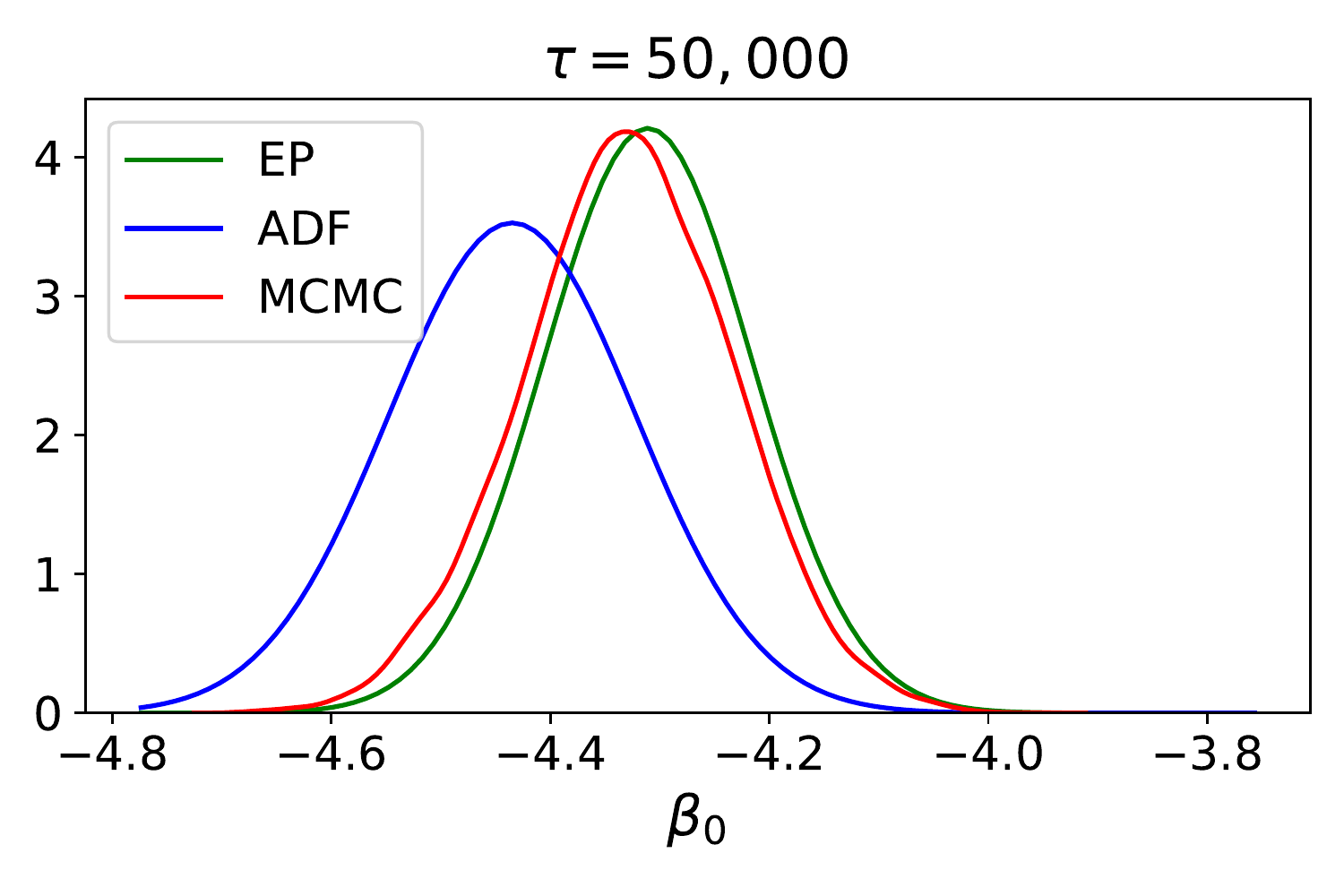}
        \label{fig:b0_50000}
    \end{subfigure}
       \begin{subfigure}[h!]{0.23\textwidth}
        \includegraphics[width=\linewidth]{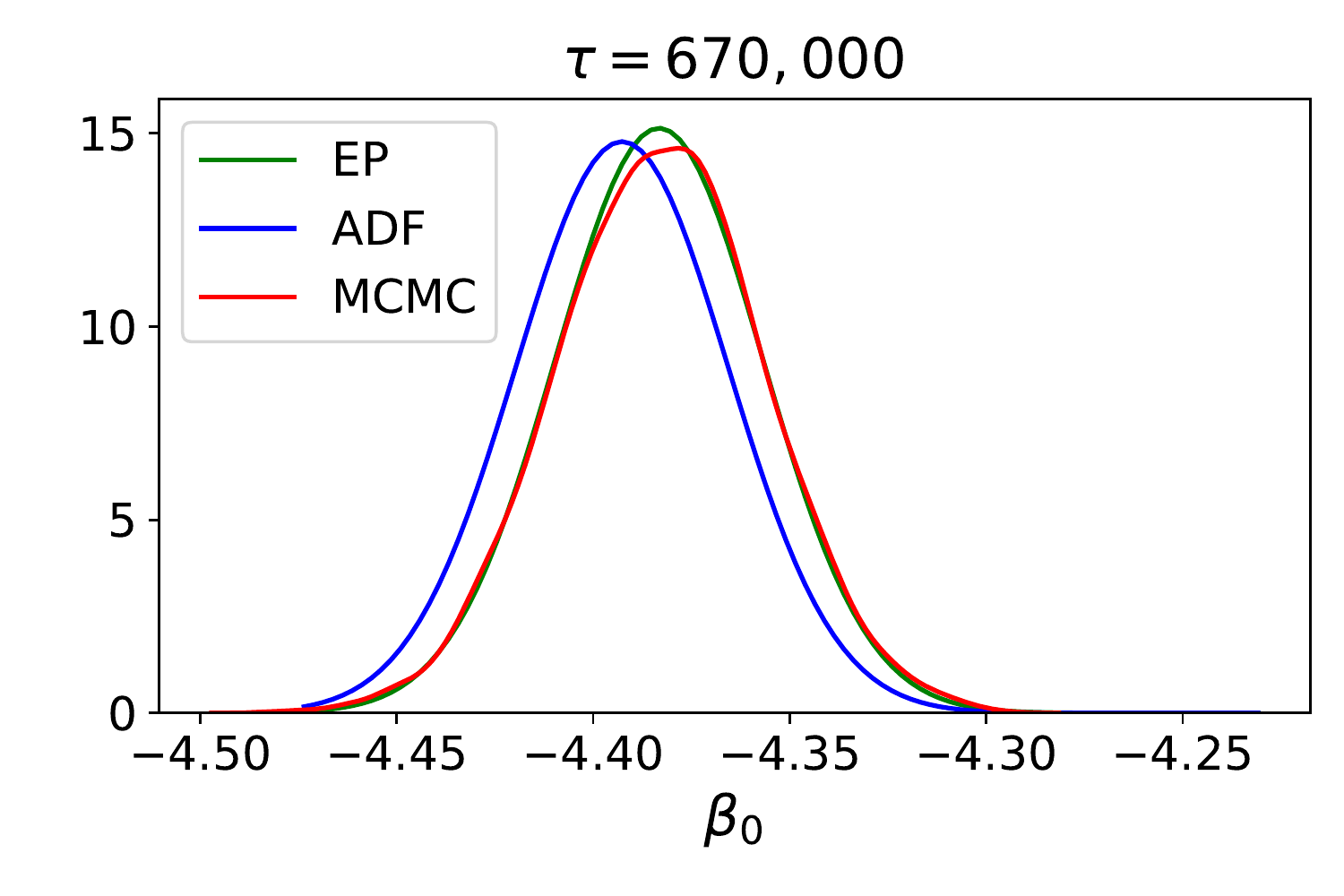}
        \label{fig:b0_full}
    \end{subfigure}
    \caption{Convergence of both EP and ADF on AT dataset. ADF is slow to
converge, motivating its combination it with
EP, which is typically much more accurate.}\label{fig:posteriors_1}
    
\end{figure}

Figure \ref{fig:posteriors_1} shows that ADF makes a very poor approximation
to the posterior unless it is trained on sufficient data. EP on the other hand
makes an accurate approximation even on the first $1,000$ data points. 

Theoretical results are given by Opper (1998) \cite{opper1998bayesian}
calculates the asymptotic convergence of ADF by showing that the inverse of the
covariance matrix approaches the fisher information matrix as $T \to \infty$.
By assuming that the difference between the change in the covariance matrix
between time points is negligible, he models its evolution as a matrix
differential equation to give asymptotic accuracy of $O(T^{-1})$ for the mean,
although no convergence rate is given for the variance. 

These empirical and theoretical considerations align with the intuition that
multiple sweeps through a dataset as in EP are expected to mitigate against
inaccuracies from sites that lead to tilted distributions that are poorly
approximated better than in a one-sweep algorithm such as ADF. Discussion by
Gelman et al.\ \cite{gelman2014expectation} is relevant in this context.
 
\subsection{Gaussian filtering}

\subsubsection{Moment matching}
As mentioned earlier, Equation~\eqref{eq:EP_objective}, which takes the form:
\begin{equation*}
q^*(\bm\theta|\bm\lambda) = \underset{\mathrm{ q \in \mathcal{Z}}}{\text{argmin}} \
	\text{KL}(t_i(\bm\theta)||q(\bm\theta|\bm\lambda)),
\end{equation*} can be solved at each iteration via moment matching 

\begin{equation*}
	\mathbb{E}_{q} \left[ \phi(\bm\theta) \right] = \mathbb{E}_{t_i}
\left[ \phi(\bm\theta) \right].
\end{equation*}
Due to the approximation being restricted to the set of Gaussian distributions, this requires only the first two moments of the tilted distribution to be
found: the mean $\bm\mu$ and the variance $\bm\Sigma$. 

Note that if all moments associated with $q(\bm\theta)$ existed and were
equal to those associated with $t(\bm\theta)$, then the
KL-divergence would be zero and the approximation would be exact.  As the
approximation used is Gaussian, it is only characterised by the first two
moments; any difference in higher moments (e.g.\ skew, kurtosis) between the
two is will lead to errors since the Gaussian lacks the flexibility to capture
these.   

In the logistic regression case, the tilted function is 
\begin{equation*}\label{eq:tilted2}
	t_i(\bm\theta) = \frac{1}{\tilde{Z}(\bm{\tilde\lambda}_i)}
p_i(\bm\theta) \prod_{j \neq i} q_j(\bm\theta|\bm\lambda_j),
\end{equation*}
where the true site distribution functions are $p_i(\bm\theta) = \sigma(y_i
\bm\theta^\top \mathbf{x}_i)$ and the approximating site functions are
$q_i(\bm\theta|\bm\lambda_i) = \mathcal{N}(\bm\theta;\bm\mu_i, \bm\Sigma_i)$. The normalisation
constant is 
 \begin{equation*}
 	\tilde{Z}(\bm{\tilde\lambda}_i) = \int p_i(\bm\theta) \prod_{j \neq i} q_j(\bm\theta|\bm\lambda_j) {\rm d} \bm\theta.
 \end{equation*}
Because of product of exponentials property (see
\S{}\ref{sec:exp_properties_prod}), the cavity can be written as $\prod_{j \neq
i} q_j(\bm\theta) = \mathcal{N}(\bm\theta;\bm\mu^{\setminus i}, \bm\Sigma^{\setminus
i})$ where the complement notation $\bm\mu^{\setminus i} :=
(\bm\mu_j)_{j \neq i}$ has been used.

Using the moments property from \S{}\ref{sec:exp_properties_moments}, the
moments of the tilted can be found via differentiation of the log partition
function as
\begin{equation*}
\mathbb{E}_{t_i} \left[ \phi(\bm\theta) \right] = \nabla_{\bm{\tilde\lambda}_i}
\log \tilde{Z}(\bm{\tilde\lambda}_i).
\end{equation*}
The remaining problem is that the natural parameters $\bm{\tilde\lambda}_i$ are
not known; if they were the moments could be found by a simple bijective
mapping. Fortunately there is a recursive formulation derived by Herbrich
(2005) \cite{herbrich2005minimising}, which enables the moments of the tilted
to be calculated as 
\begin{equation}\label{eq:recursive_moment_update1}
 \mathbb{E}_{t_i} \left[ \phi(\bm\theta) \right]
 =  \mathbb{E}_{q^{\setminus i }} \left[ \phi(\bm\theta) \right]+ \nabla_{\bm\lambda^{\setminus i}} \log \tilde{Z}(\bm{\tilde\lambda}), 
 \end{equation}
meaning that the cavity natural parameters $\bm\lambda^{\setminus i}$ are needed
instead of the tilted natural parameters $\bm{\tilde\lambda}_i$.
 
After some calculations, \eqref{eq:recursive_moment_update1} can be used to
give an iterative update formula for the first and second Gaussian moments of
the global approximation:
\begin{equation}
\label{eq:moment_update1}
\bm{\mu} = \bm{\mu}^{\setminus i } + \bm\Sigma^{\setminus i } \bm\alpha_i
,\quad
\bm{\Sigma}  = \bm{\Sigma}^{\setminus i } - \bm{\Sigma}^{\setminus i
}\left(\bm{\alpha}_i \bm{\alpha}_i^\top - 2 \mathbf{B}_i
\right)\bm{\Sigma}^{\setminus i },
\end{equation}
where 
\begin{equation}\bm\alpha_i = \nabla_{\bm\mu_i} \log \tilde{Z}(\tilde{\bm\lambda}_i) \in
\mathbb{R}^D \hspace{0.23cm} \text{and} \hspace{0.23cm} \mathbf{B}_i = \nabla_{\bm\Sigma_i} \log
\tilde{Z}(\tilde{\bm\lambda}_i) \in \mathbb{R}^{D \times D}.\end{equation}


 \subsubsection{Linear subspace property}

The normalising constant to the tilted distribution
$\tilde{Z}(\tilde{\bm\lambda}_i)$ is, in general, intractable -- its moments can
be evaluated via numerical quadrature or MCMC, however in this work an approximation
described by MacKay (1992) \cite{mackay1992evidence} is used.  In MacKay's estimation
framework, known as the `evidence framework' or `moderated output', the normalising constant is approximated as
 \begin{equation}
\tilde{Z}(\bm{\tilde\lambda}_i) = \int \sigma(y_i \bm\theta^\top \mathbf{x}_i)
\mathcal{N}(\bm\theta;\bm\mu^{\setminus i}, \bm\Sigma^{\setminus i}) {\rm d} \bm\theta
\approx \sigma(\kappa(s_i^2) a_i ) \label{eq:moderated_output_approximation},
 \end{equation}
where
\begin{equation*}
\kappa (s_i^2) = \big( 1 + (\pi s_i^2/8) \big)^{-1/2} , \quad
s_i^2 = \mathbf{x}_i^\top \bm\Sigma^{\setminus i} \mathbf{x}_i,
\quad a_i = \mathbf{x}_i^\top \bm\mu^{\setminus i}.
\end{equation*}
Using the moderated output given in \eqref{eq:moderated_output_approximation},
$ \bm{\alpha}_i$ and $\mathbf{B}_i$ are expressed as
\begin{equation}
 \begin{aligned} \label{eq:alphaB}
\bm{\alpha}_i = \nabla_{\bm\mu^{\setminus i}} \log \tilde{Z}(\bm{\tilde\lambda}_i) & =  \bm\rho_i \frac{\sigma'(\bm\rho_i \bm{\mu}^{\setminus i})}{\sigma(\bm\rho_i \bm{\mu}^{\setminus i})},\\
	\mathbf{B}_i = \nabla_{\bm\Sigma^{\setminus i}} \log \tilde{Z}(\bm{\tilde\lambda}_i) & = -\frac{\pi  }{16 \kappa^2} \mathbf{x}_i \mathbf{x}_i^T \bm{\mu}^{\setminus i} \bm{\alpha}_i^T,
\end{aligned}
\end{equation}
where
\begin{equation*}
\bm\rho_i = y_i \kappa_i(s^2)   \mathbf{x}_i^T,
\ \sigma(z) = \frac{1}{1 + \exp(-z)},
\ \sigma'(z) = \frac{\exp(-z)}{(1 + \exp(-z))^2}.
\end{equation*}

\subsubsection{Filtering algorithm}

Using the results \eqref{eq:alphaB} along with the iterative update equations
\eqref{eq:moment_update1}, Gaussian approximations to the posterior in Bayesian
logistic regession can be computed for both EP and ADF. These are shown in Algorithms
\ref{alg:EP_calculations} and \ref{alg:ADF_calculations} respectively. 

 \begin{algorithm}
 Initialise the global approximation $q(\bm\theta) = \mathcal{N}(\bm\mu_0, \bm\Sigma_0)$\\
 \While{not converged}
 {
 \For{$ i = 1 \dots T$}
 {1. Form the cavity expected moments: \\
 	\hspace{0.27cm} $\bm{\Sigma}^{\setminus i } = \left(  \bm{\Sigma}^{-1} - (\bm{\Sigma}_i)^{-1} \right)^{-1}$\\
 	\hspace{0.27cm} $\bm{\mu}^{\setminus i} = \bm{\Sigma}^{\setminus i } \left( \bm{\Sigma}^{-1}\bm{\mu} - \bm{\Sigma}_i^{-1}\bm{\mu}_i \right) $\\
 2. Project the moments of the tilted distribution onto the global approximation: \\
 	\hspace{0.27cm} $\bm{\mu}  = \bm{\mu}^{\setminus i } + \bm\Sigma^{\setminus i } \bm\alpha_i$\\
 	\hspace{0.27cm} $\bm{\Sigma}  = \bm{\Sigma}^{\setminus i } - \bm{\Sigma}^{\setminus i }\left(\bm{\alpha}_i \bm{\alpha}_i^\top - 2 \mathbf{B}_i \right)\bm{\Sigma}^{\setminus i }$ \\
 3. Update the expected moments of site $i$:\\
 	\hspace{0.27cm} $\bm{\Sigma}_i = \left(  \bm{\Sigma}^{-1} - (\bm{\Sigma}^{\setminus i })^{-1} \right)^{-1}$\\
 	\hspace{0.27cm} $\bm{\mu}_i = \bm{\Sigma}_i \left( \bm{\Sigma}^{-1}\bm{\mu} - (\bm{\Sigma}^{\setminus i })^{-1}\bm{\mu}^{\setminus i } \right) $\\
 } }
\caption{Gaussian Expectation Propagation}\label{alg:EP_calculations}
\end{algorithm}

 \begin{algorithm}
Initialise the prior distribution $q_0(\bm\theta) = \mathcal{N}(\bm\mu_0, \bm\Sigma_0)$\\
 \For{$ i = 1 \dots T$}
 {
 1. The cavity distribution is simply the approximation made at the previous iteration:\\
 	\hspace{0.27cm} $\bm{\mu}^{\setminus i }  =  \bm{\mu}_{i-1}$\\
 	\hspace{0.27cm} $\bm{\Sigma}^{\setminus i }  =  \bm{\Sigma}_{i-1}$\\
 2. Project the moments of the tilted distribution onto the global approximation: \\
 	\hspace{0.27cm} $\bm{\mu}  = \bm{\mu}^{\setminus i } + \bm{\Sigma}^{\setminus i } \bm{\alpha}_i$\\
 	\hspace{0.27cm} $\bm{\Sigma}  = \bm{\Sigma}^{\setminus i } - \bm{\Sigma}^{\setminus i }\left(\bm{\alpha}_i \bm{\alpha}_i^\top - 2 \mathbf{B}_i \right)\bm{\Sigma}^{\setminus i }$ \\

 } 
\caption{Gaussian Density Filtering}\label{alg:ADF_calculations}
\end{algorithm} 

\subsection{Combining methods in FAB-COST}

An outline is now given of how the considerations above lead us to the FAB-COST approach
to recommendation systems, which will turn out to provide improved performance
on real data for well-controlled computational effort.

  \begin{figure}[h!]
     \begin{subfigure}[h!]{0.23\textwidth}
        \includegraphics[width=\linewidth]{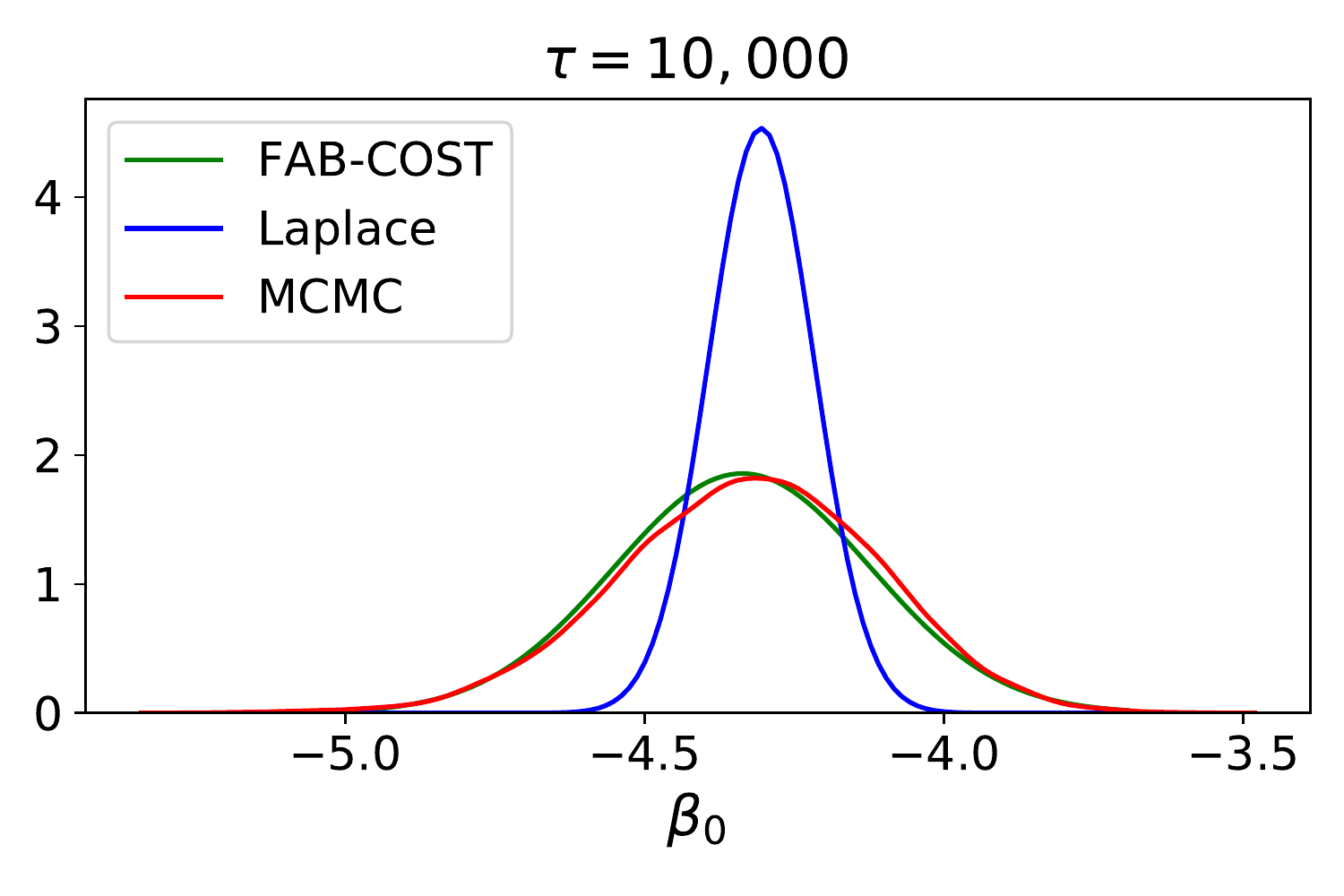}
        \label{fig:b0_1000}
    \end{subfigure}
    \begin{subfigure}[h!]{0.23\textwidth}
        \includegraphics[width=\linewidth]{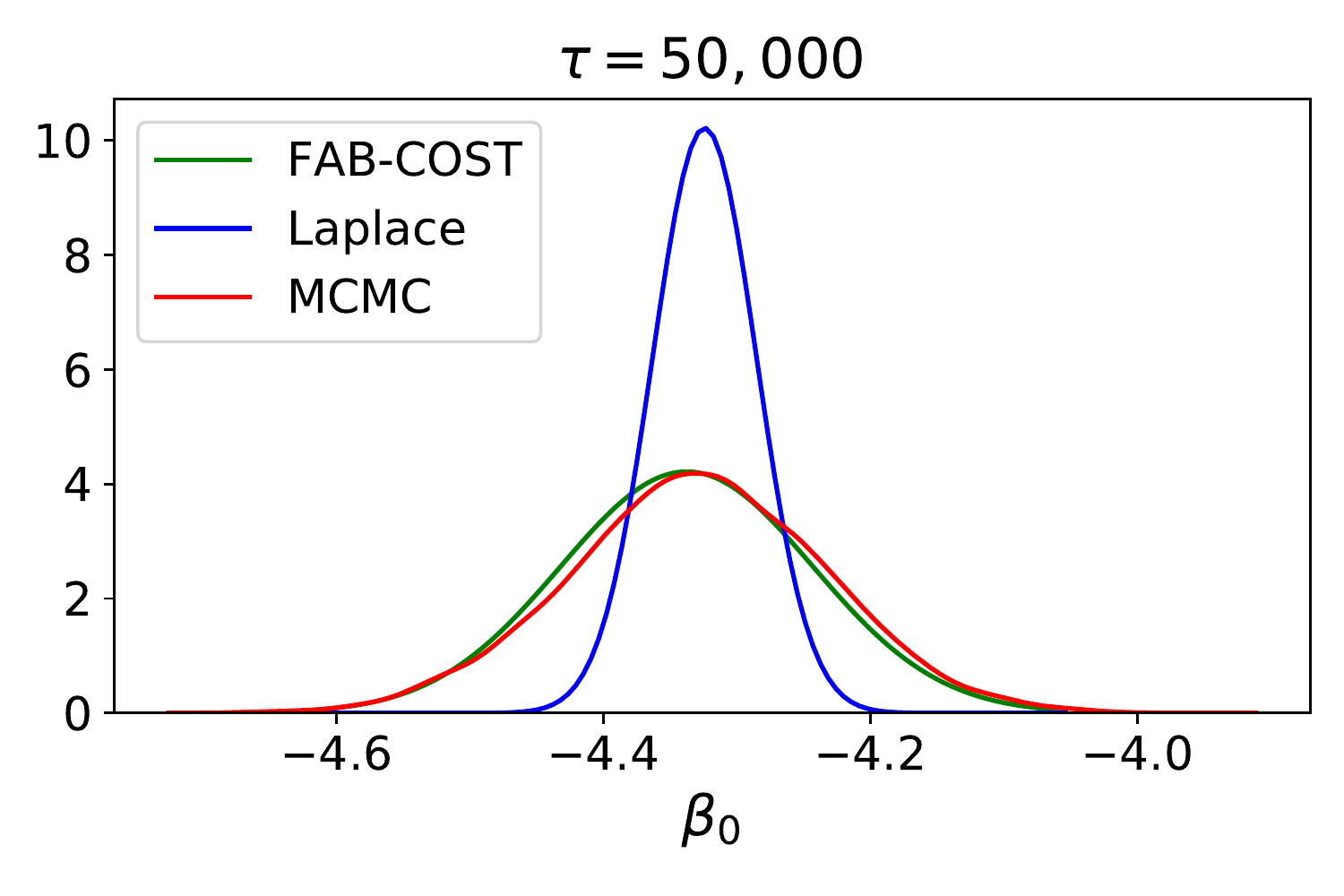}
        \label{fig:b0_10000}
    \end{subfigure}
    \begin{subfigure}[h!]{0.23\textwidth}
        \includegraphics[width=\linewidth]{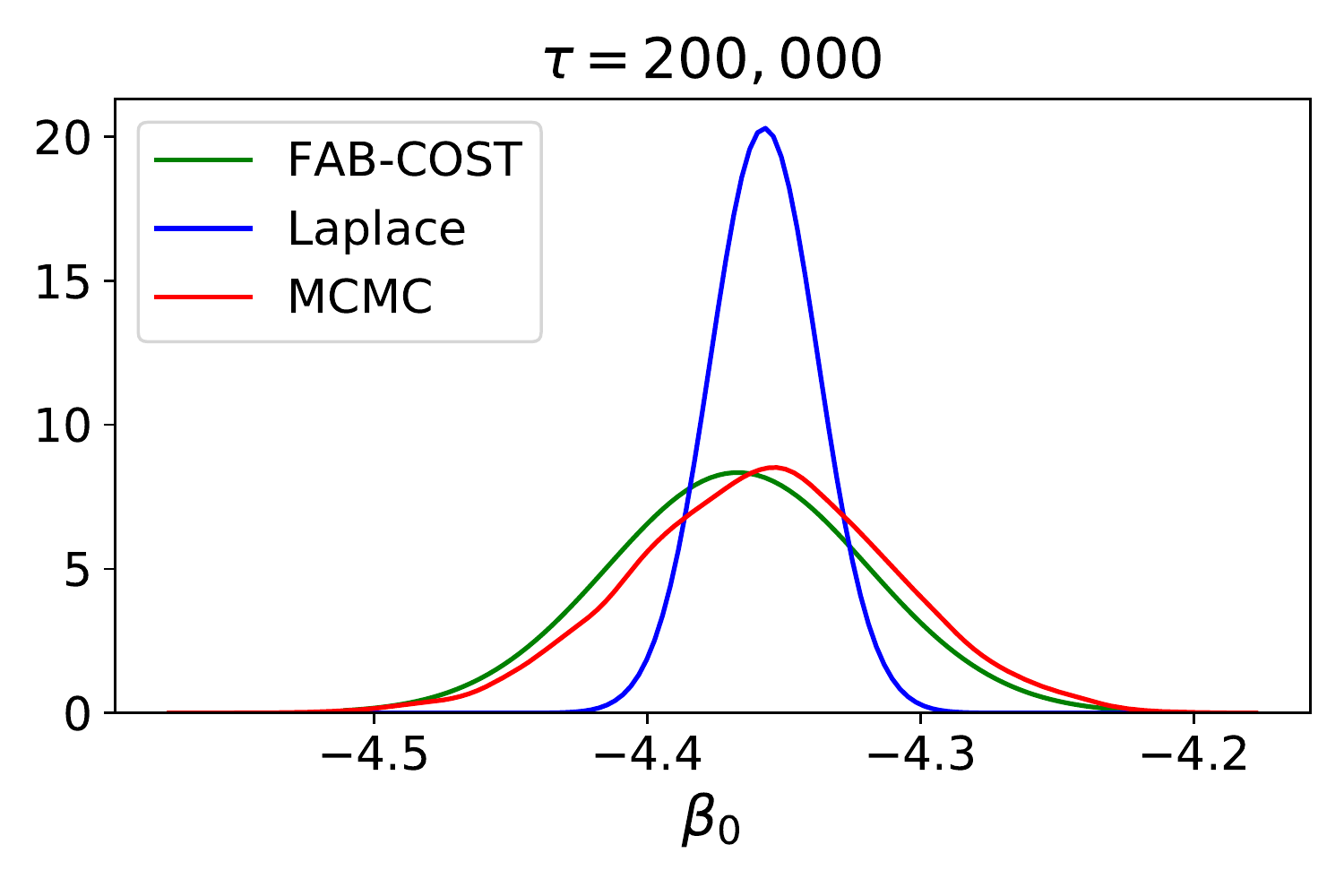}
        \label{fig:b0_50000}
    \end{subfigure}
       \begin{subfigure}[h!]{0.23\textwidth}
        \includegraphics[width=\linewidth]{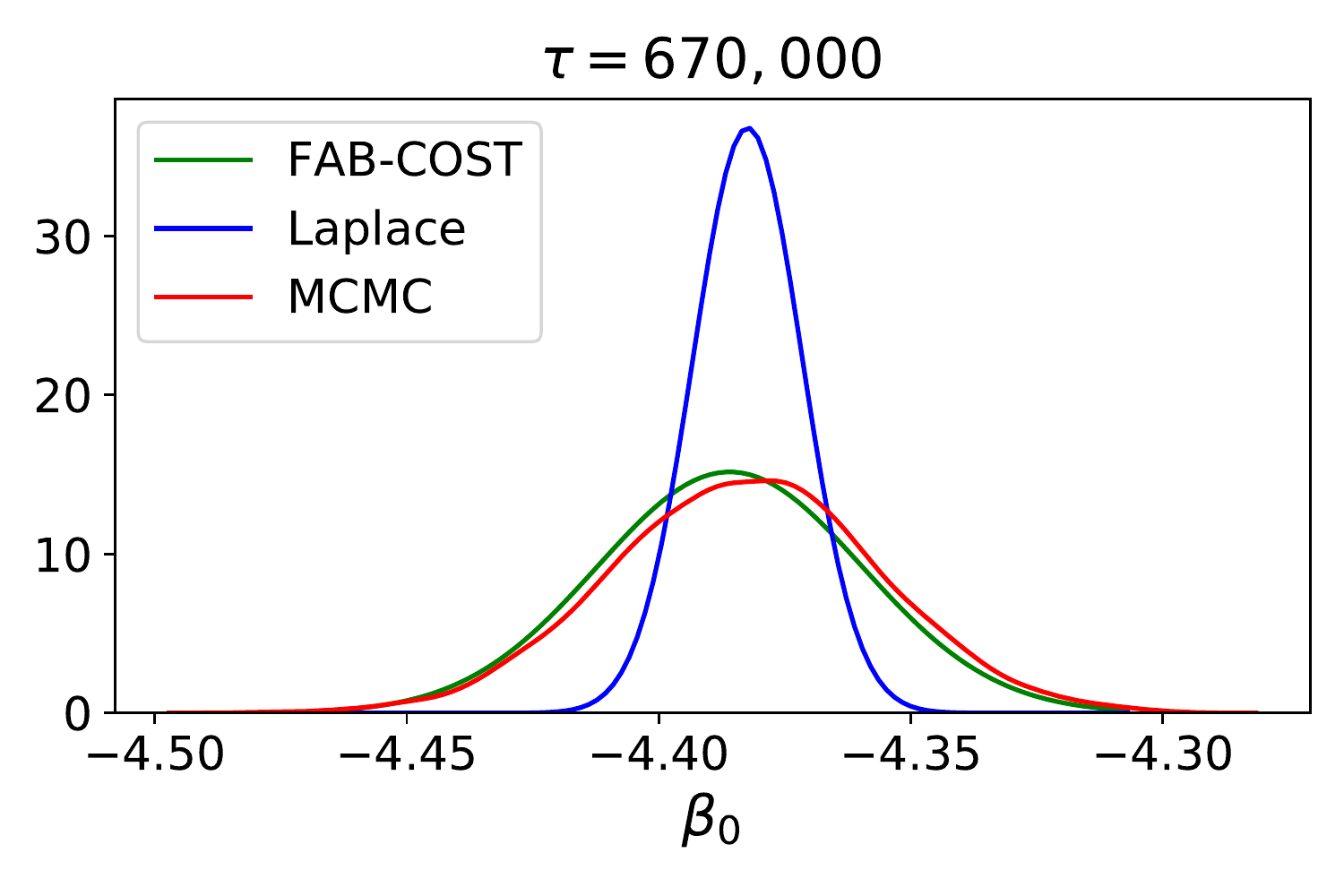}
        \label{fig:b0_full}
    \end{subfigure}
    \caption{FAB-COST vs Laplace. The Laplace approximation fails to capture
the true variance, meaning that in the bandit setting a failure to
balance the exploration-exploitation tradeoff is expected.}\label{fig:posteriors_2}
    
\end{figure}

\subsubsection{Data}

Auto Trader PLC is the UK's largest digital automotive marketplace, and the `AT' dataset was constructed from the website autotrader.co.uk. This consists
of a day's `featured listing' user click data, totalling at $T = 678,446$
impressions.  After a user makes a search for a car they are presented with a
single `featured listing' which appears at the top of the search result,
meaning that there is no positional bias. The user then proceeds to either click
or not click on the presented advert.  Many dozens of covariates for each
advert are available, which was reduced to the $15$ most important using a
random forest algorithm for feature selection, although other possibilities for
feature selection can be used \cite{Hastie:2009}.

\subsubsection{Computational Cost}\label{sec:computational Cost}

The structure of Algorithms \ref{alg:EP_calculations} and
\ref{alg:ADF_calculations} makes clear that one iteration ADF is expected to be
computationally cheaper than one iteration of EP. This is because, at each
iteration, EP requires the expensive matrix inversion required to map between
the natural and moment parameters. Although both matrix multiplication and
matrix inversion come at cubic computational complexity $O(D^3)$, the
pre-multiplication constant for ADF is much smaller. Calculations show that
per site, EP's flop count of $\frac{38}{3} D^3 + O(D^2)$  is over three times
greater than that of ADF's flop count of $4 D^3 + O(D^2)$. Due to matrix
multiplication being easily parallelised (which Python's Numpy library
exploits), as well as EP requiring multiple sweeps over the data set, we
recorded a $75$-fold speed-up when using ADF in batch over EP.

Now consider the number of iterations involved in an online learning context.
Let $\tau$ be the number of data points used to make the last posterior
approximation and $m$ is the number of data points arriving since the last
posterior approximation. EP requires the entire data set up to the current
iteration to make a posterior approximation leading to a computational cost of
$O((\tau+m) D^3)$ compared to ADFs $O( m D^3)$. For $m\ll \tau$ this is going
to result in a dramatic increase in computational cost choosing EP over ADF. 

\begin{figure}[h!]
     \begin{subfigure}[h!]{0.23\textwidth}
        \includegraphics[width=\linewidth]{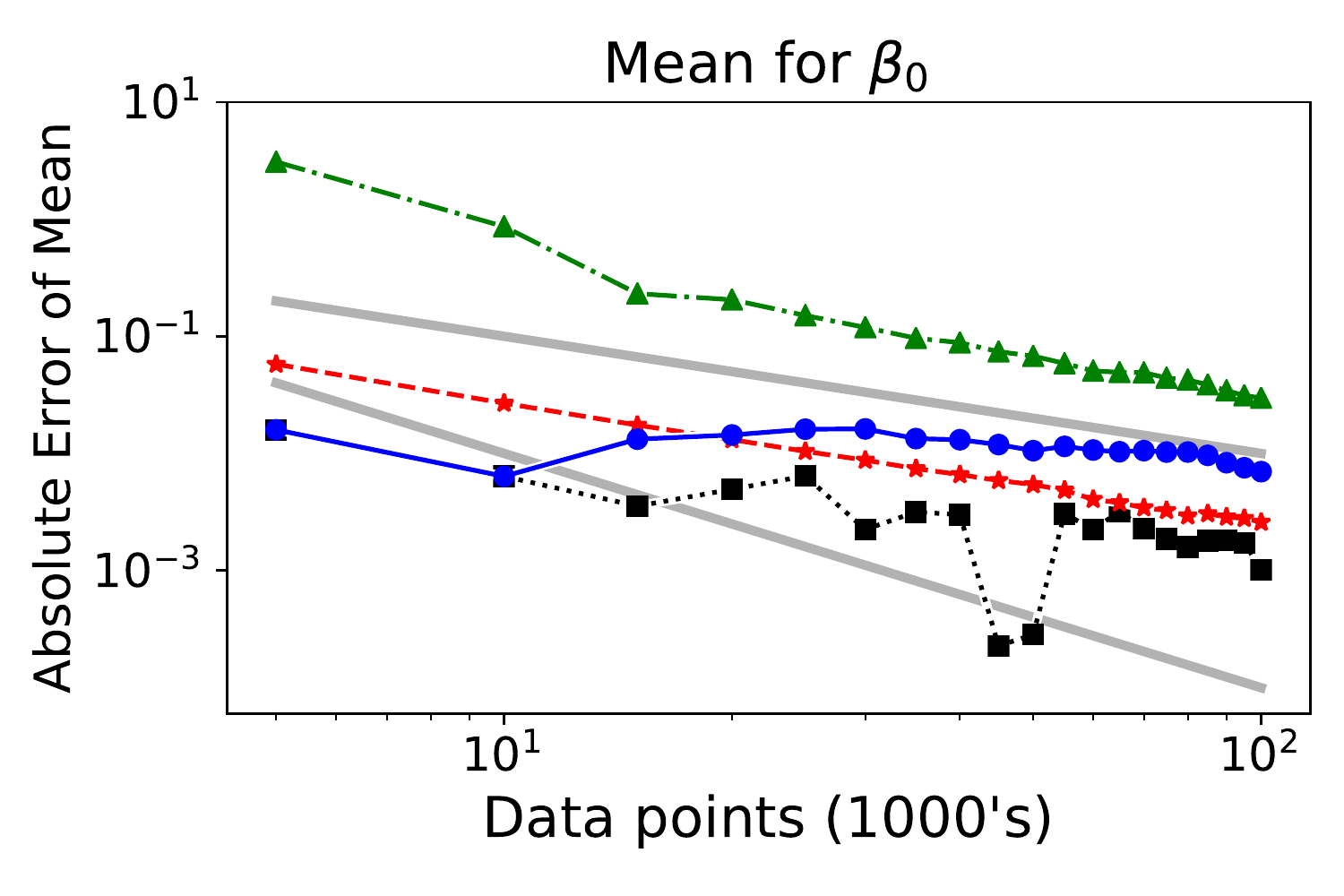}
    \end{subfigure}
    \begin{subfigure}[h!]{0.23\textwidth}
        \includegraphics[width=\linewidth]{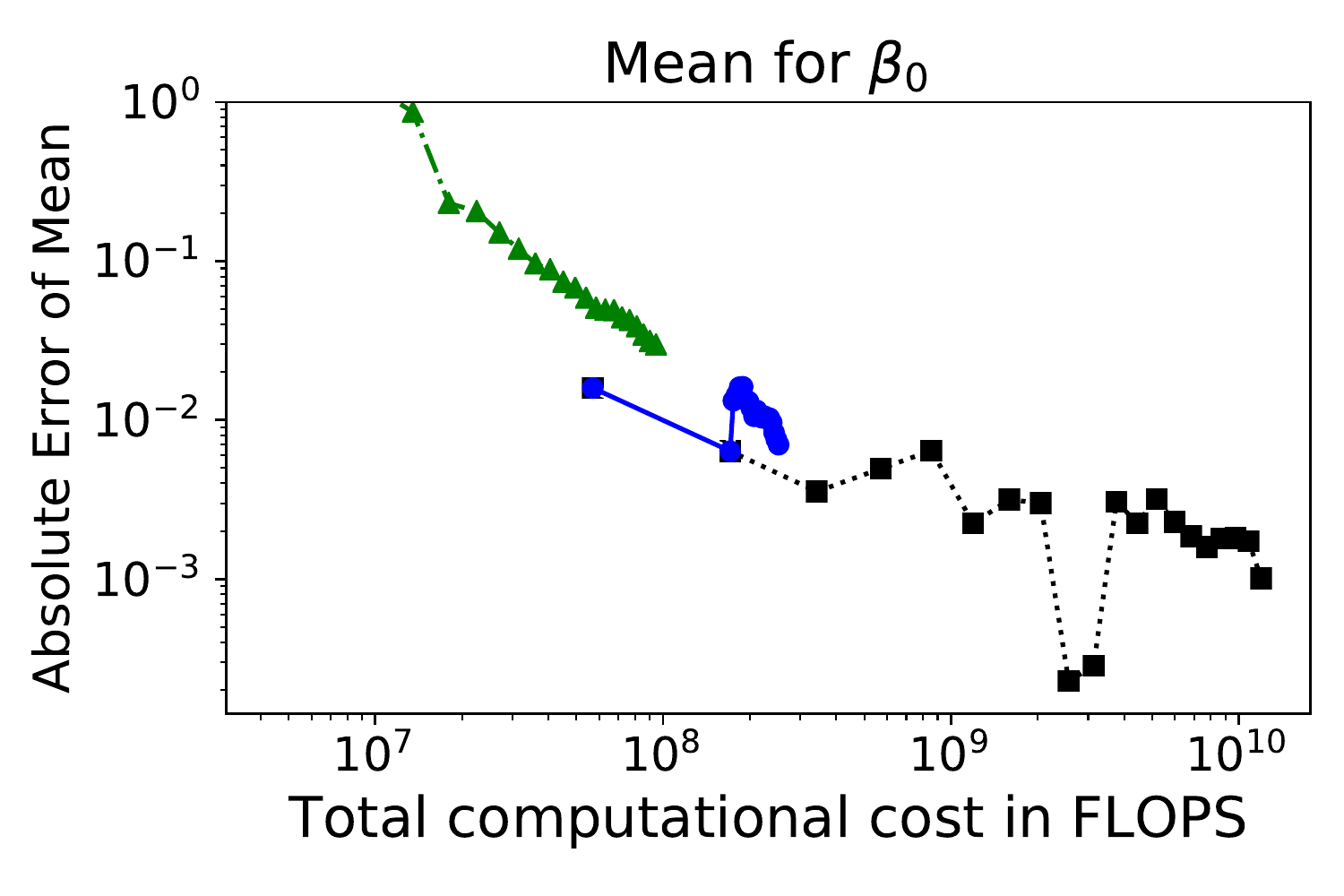}
    \end{subfigure}
    \begin{subfigure}[h!]{0.23\textwidth}
        \includegraphics[width=\linewidth]{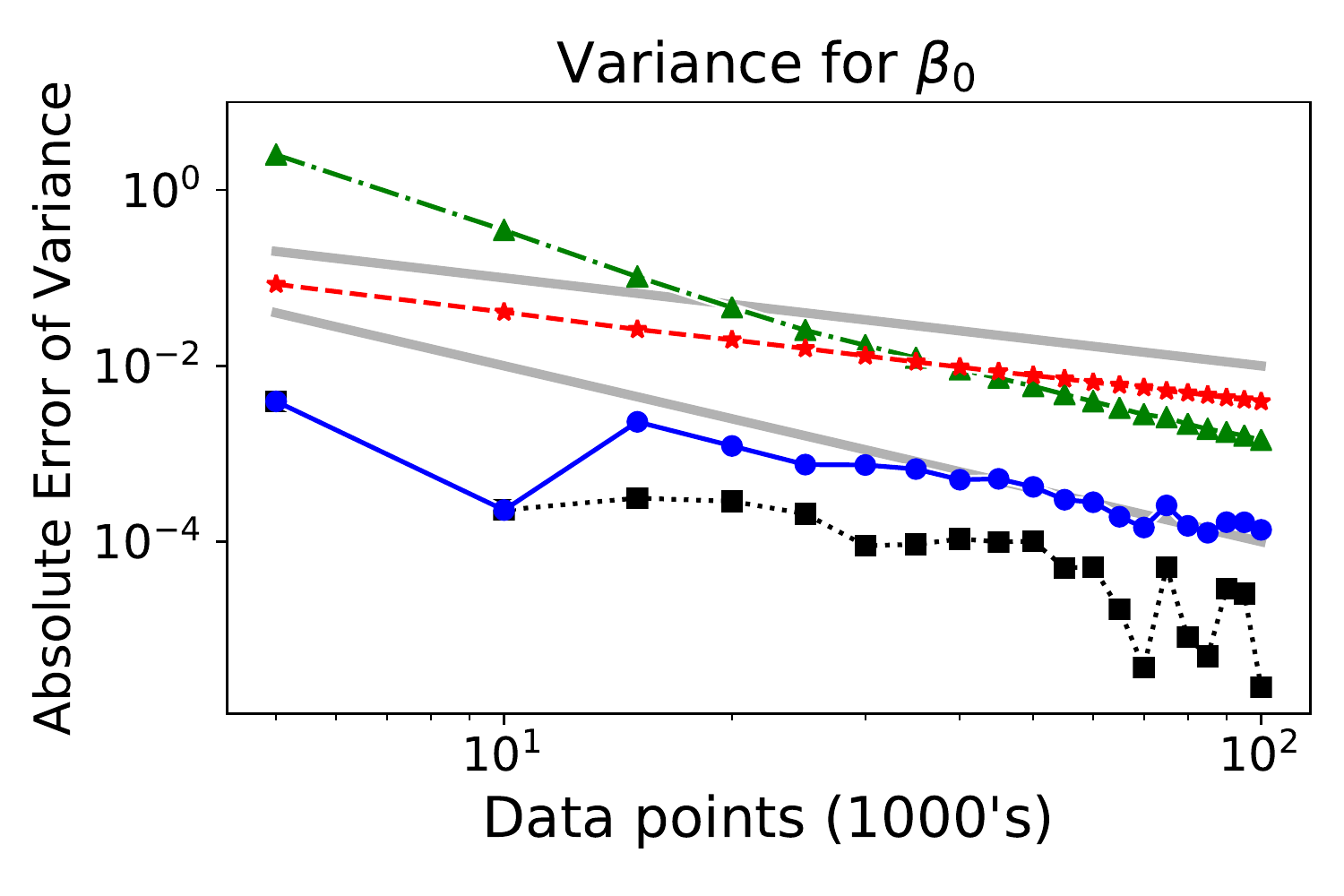}
    \end{subfigure}
       \begin{subfigure}[h!]{0.23\textwidth}
        \includegraphics[width=\linewidth]{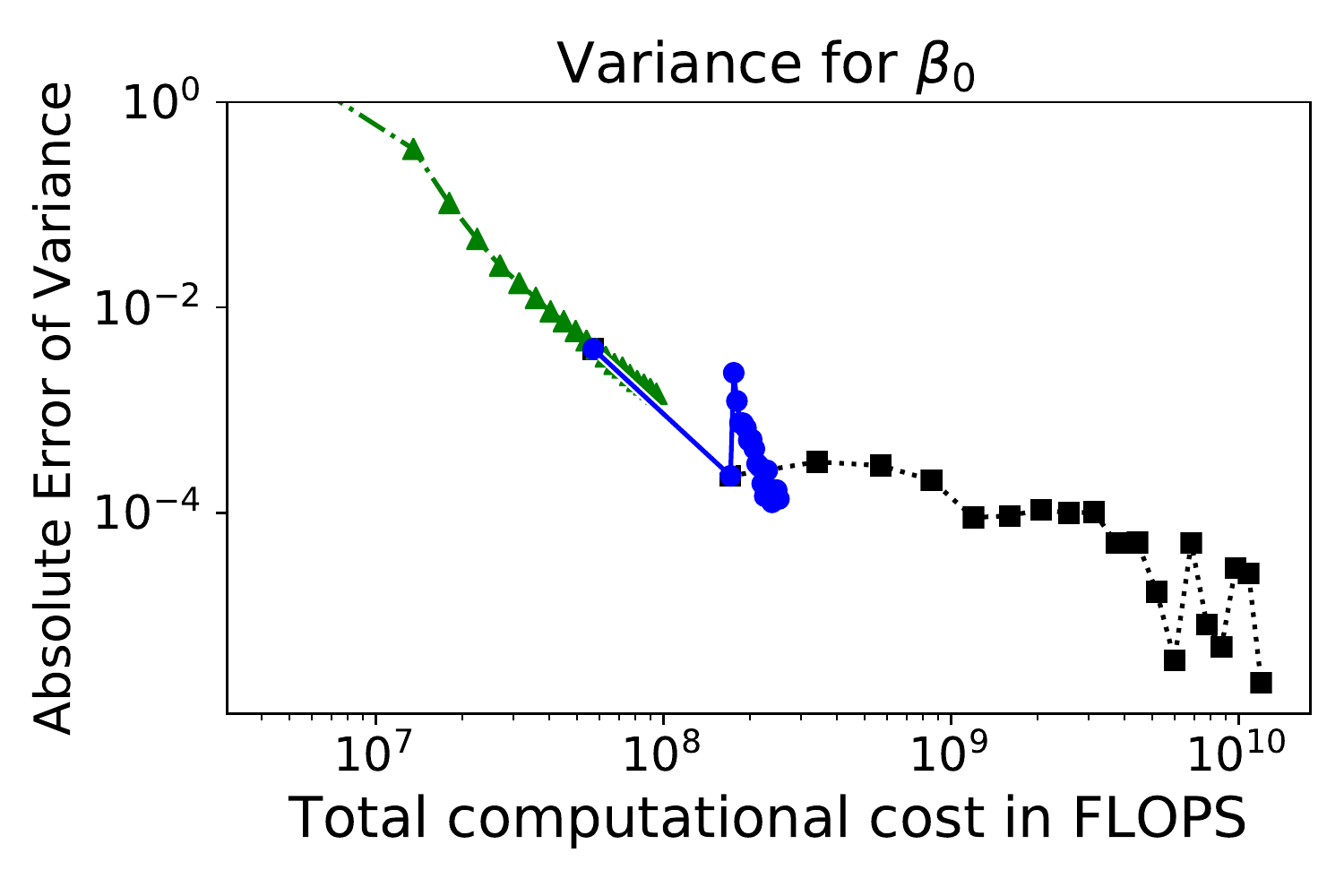}
    \end{subfigure}
         \begin{subfigure}[h!]{0.23\textwidth}
        \includegraphics[width=\linewidth]{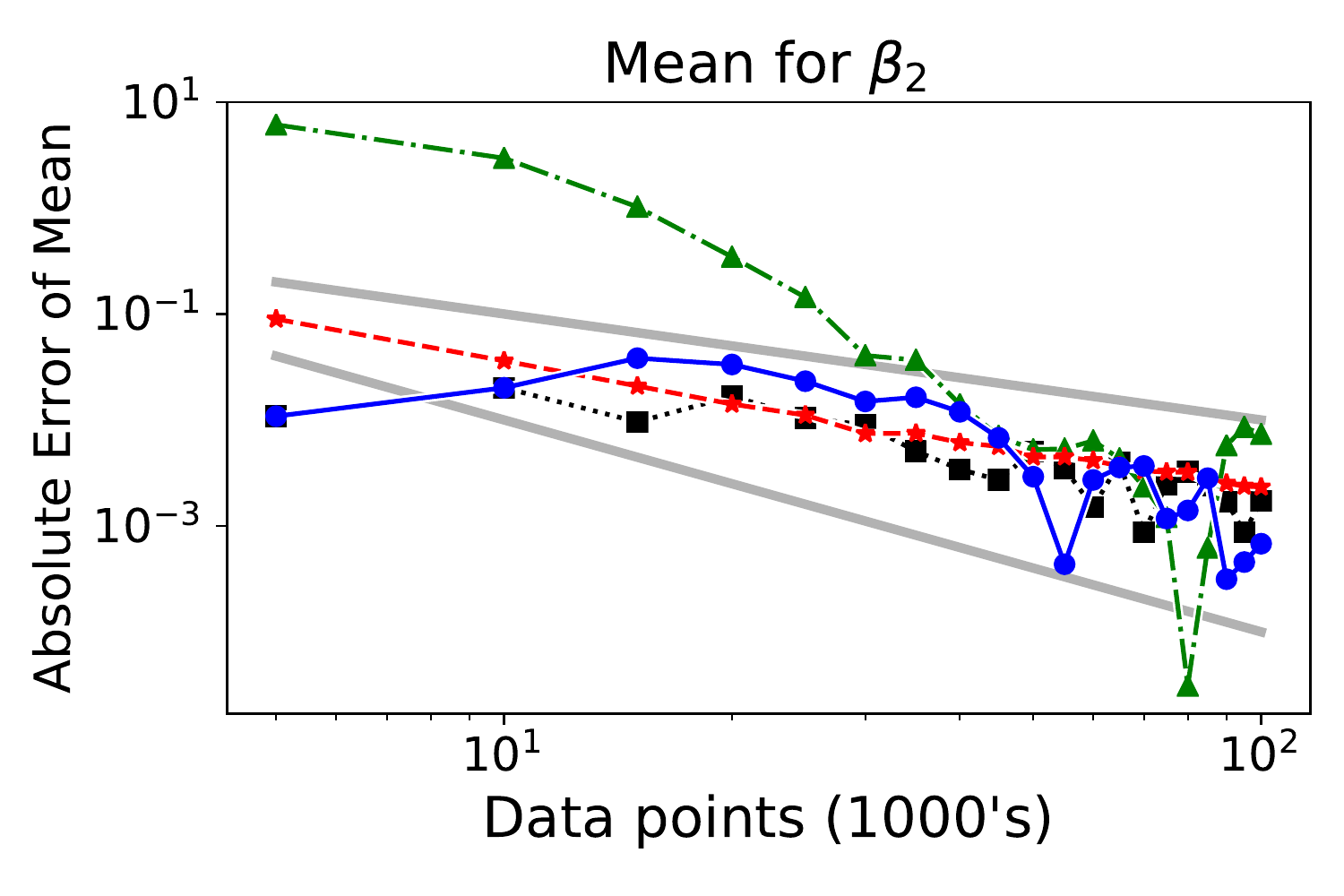}
    \end{subfigure}
    \begin{subfigure}[h!]{0.23\textwidth}
        \includegraphics[width=\linewidth]{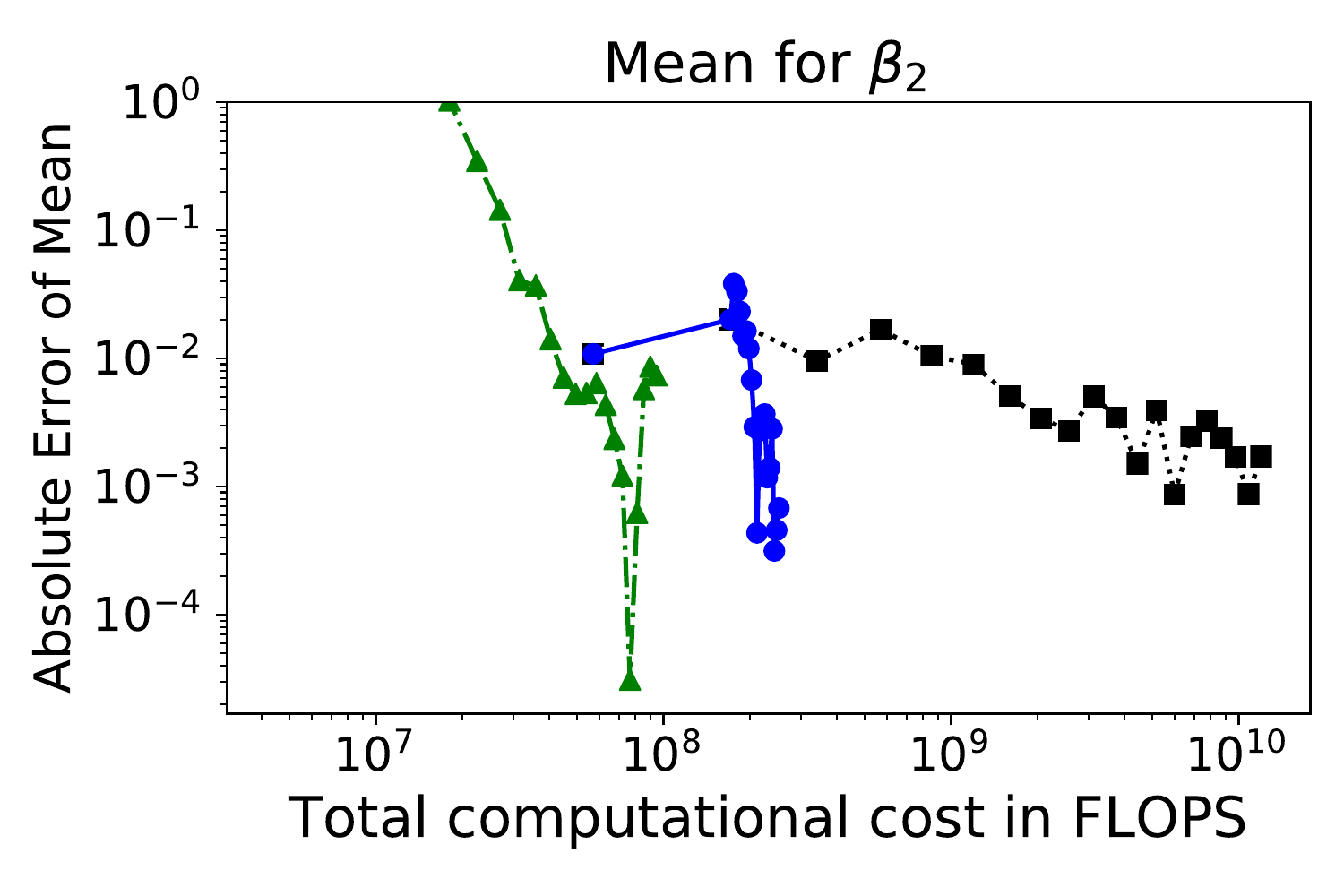}
    \end{subfigure}
    \begin{subfigure}[h!]{0.23\textwidth}
        \includegraphics[width=\linewidth]{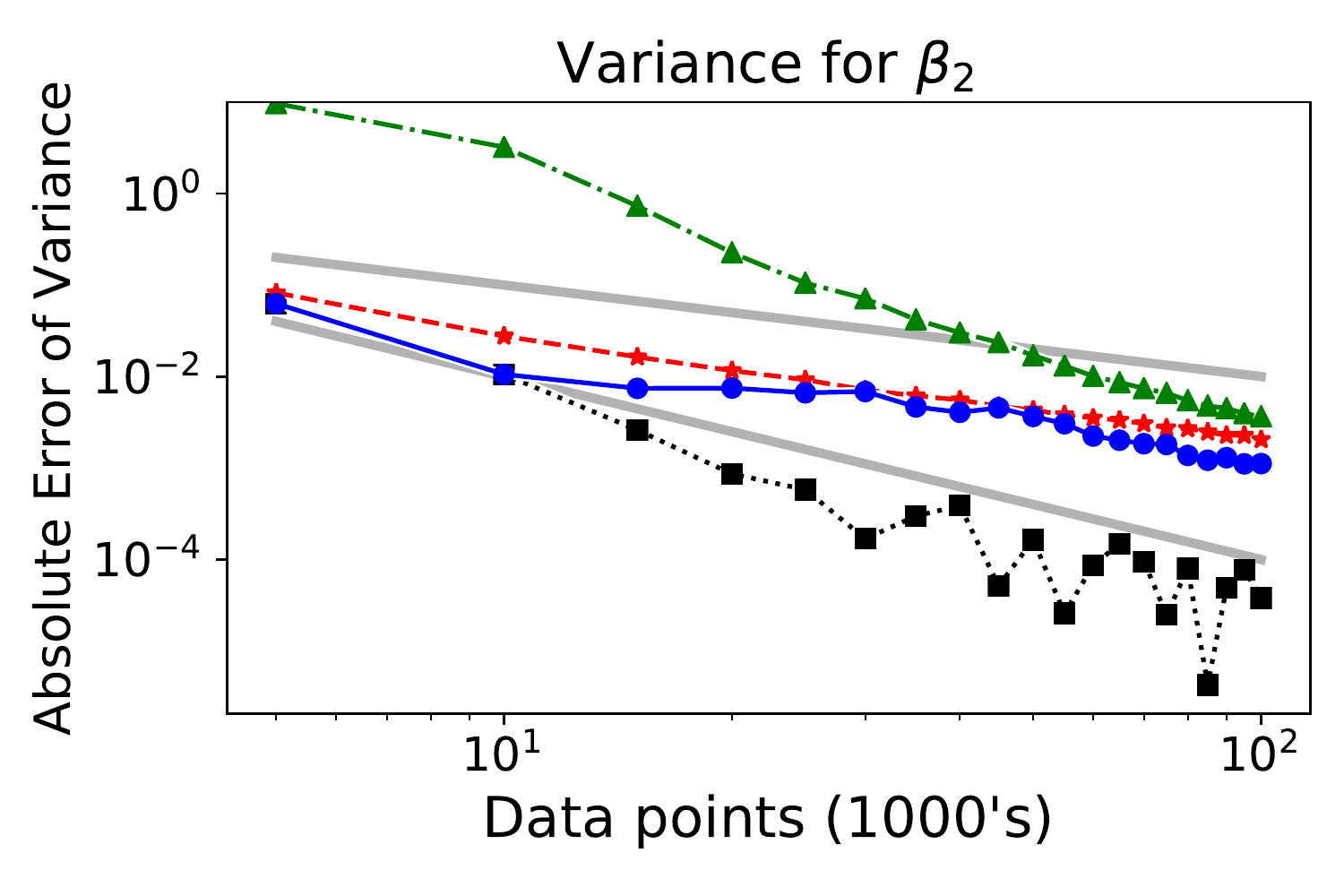}
    \end{subfigure}
       \begin{subfigure}[h!]{0.23\textwidth}
        \includegraphics[width=\linewidth]{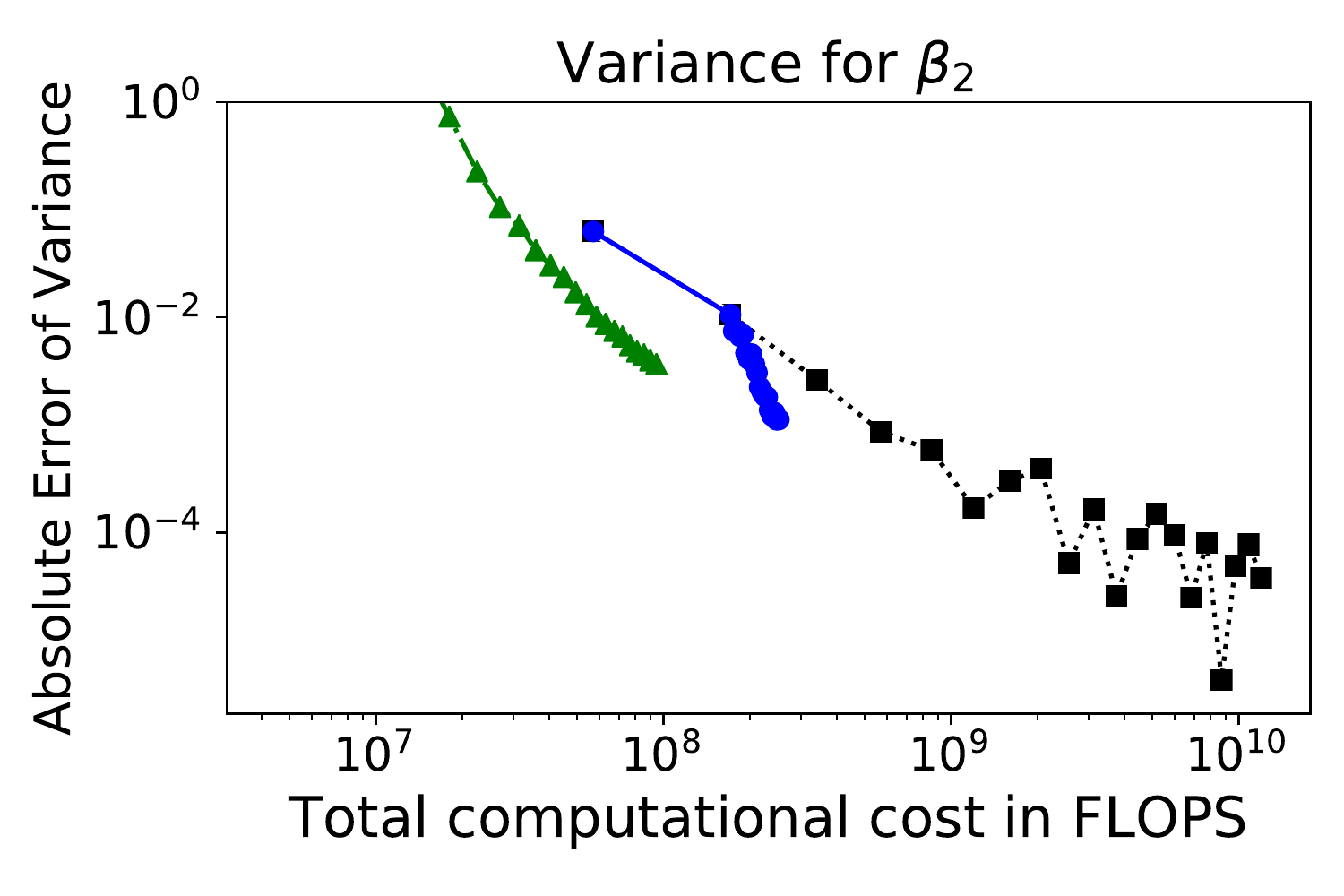}
    \end{subfigure}
        \begin{subfigure}[l]{0.08\textwidth}
        \includegraphics[width=\linewidth]{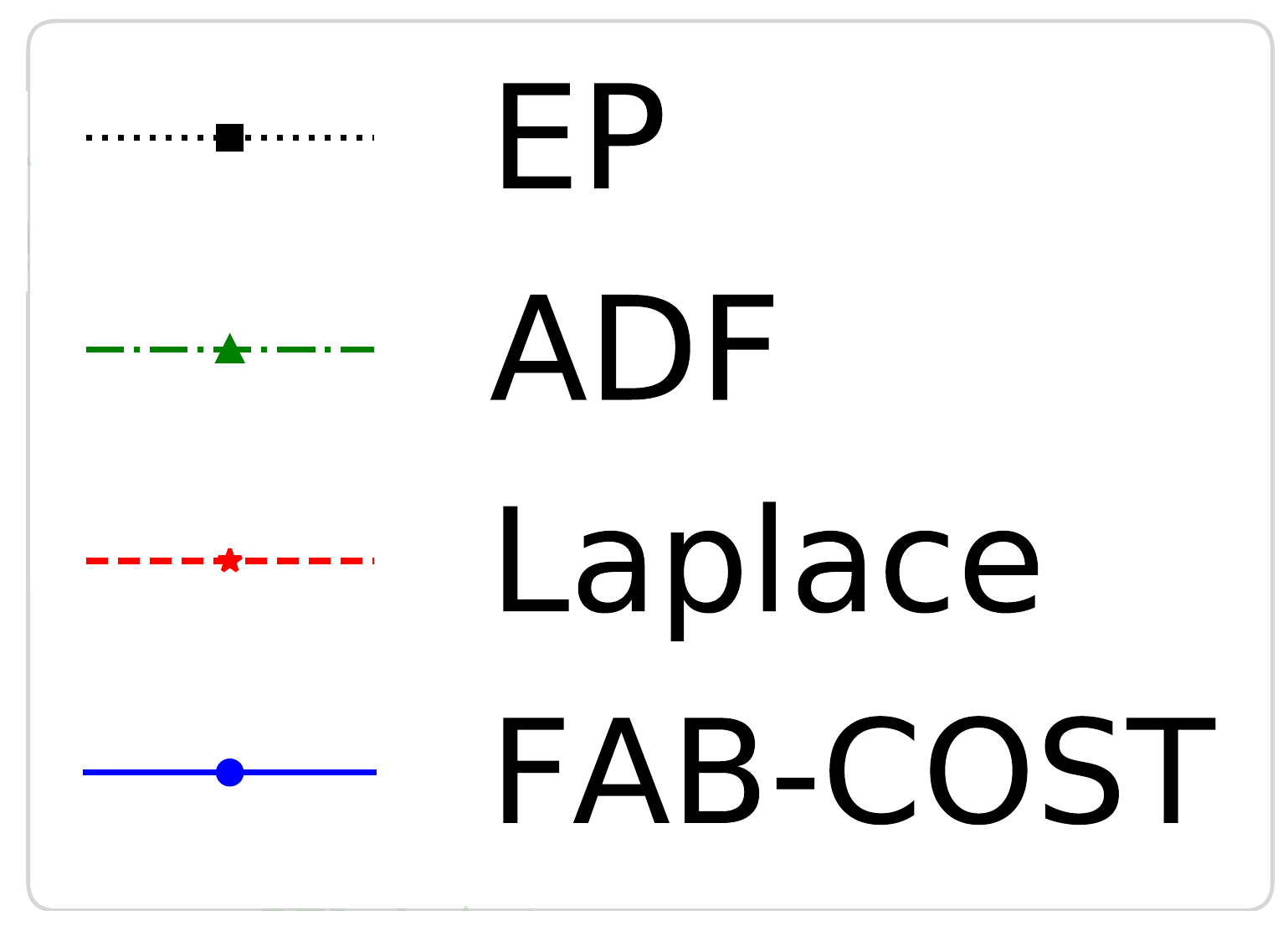}
    \end{subfigure}
    \caption{The left column shows log-log plots of the mean and variance error
for the Laplace approximation, ADF, EP and FAB-COST where the No U-Turn sampler
was used to establish the ground truth. The grey lines show asymptotic error of
$O(T^{-1})$ and $O(T^{-2})$ for reference. The right column shows the asymptotic
accuracy computational cost trade-off between the methods as discussed in
section \ref{sec:computational Cost}.}\label{fig:moment_convergence_rates}
    
\end{figure}

The computational cost vs accuracy trade-off is shown in the right column of
Figure \ref{fig:moment_convergence_rates}. The requirement for EP to be ran in
batch is what causes the significant increase in the FLOPs use over the learning
process. In these plots, an EP update is performed every $5,000$ impressions
($m=5,000$) rather than in a true sequential manner. This batch size was chosen
purely for illustrative purposes; an EP update would have to be done more
frequently early on in a bandit setting. If these batch sizes decreased then
the computational cost would increase as a result.

\subsubsection{Accuracy}

As discussed above, MCMC can be used to provide an arbitrarily accurate
representation of the posterior, although the computational effort involved in
doing this is prohibitive in an online context. To assess accuracy for the
purposes of this study, however, the No-U-Turn sampler (NUTS) was
implemented using PyMC3 \cite{Salvatier:2016}, and takes the output of this as the
ground truth of the posterior distributions we are trying to approximate. By
measuring the absolute error between the moments of the ground truth and each
inference procedure, an empirical estimate for the asymptotic
convergence rate of each can be computed by measuring the gradient of a log-log plot (see the
left column of Figure \ref{fig:moment_convergence_rates}).

The results show that EP gives better asymptotic accuracy to the posterior than
other methods, as would be expected from the theoretical results discussed
above, although for the real dataset considered, a simple power law in $T$ is not
observed. In general it is clear that the main improvements for EP and ADF over
Laplace come from estimation of the variance, as Figure \ref{fig:posteriors_2}
shows more explicitly. This to be particularly important for
recommendation systems, where balancing the explore-exploit tradeoff is
crucial.

Also as expected, once data availability becomes large, one can switch to ADF as
proposed in the FAB-COST algorithm without significant loss of accuracy.
Although EP can be used periodically to update the posterior approximation in
the FAB-COST algorithm, only two EP updates were performed in the simulations above
(at $100$ and $10,000$ iterations) although the posterior accuracy would be improved with
more frequent EP updates. Theses updates at $100$ and $10,000$ iterations out of the
$T = 670,000$ sized data set shows that if given a reasonable prior, ADF
achieves good accuracy. 
 
%

\subsubsection{Recommendation system and performance}

Algorithm \ref{alg:FAB-COST} shows the pseudocode for FAB-COST. This takes the
logistic bandit (shown in Algorithm \ref{algorithm:posterior_sampling}) and
adds moment updating using both EP and ADF. In the experiments, two EP updates were chosen at $E = 100$ and $E = 10,000$ (although this can be performed as
often as is computationally feasible) with ADF sequentially updating the posterior approximation at each
iteration for the remainder of the simulation. These EP updates were chosen
to show that even with few updates, FAB-COST is superior to the Laplace
approximation for both posterior approximation accuracy and reward that is achieved in the
bandit setting. 

\begin{figure}
    \begin{subfigure}[h!]{0.23\textwidth}
        \includegraphics[width=\linewidth]{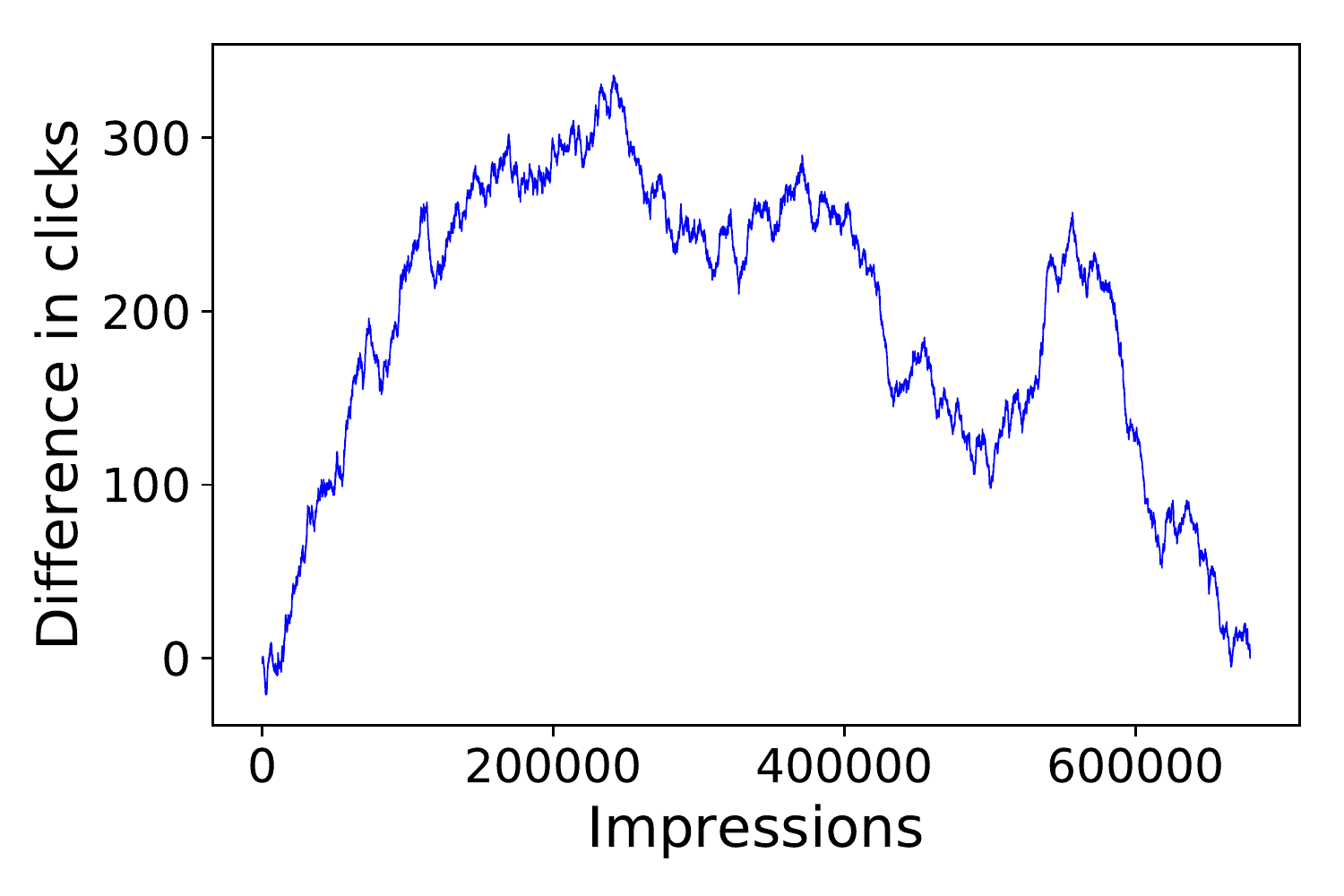}
        \label{fig:FC_difference_in_clicks}
    \end{subfigure}
       \begin{subfigure}[h!]{0.23\textwidth}
        \includegraphics[width=\linewidth]{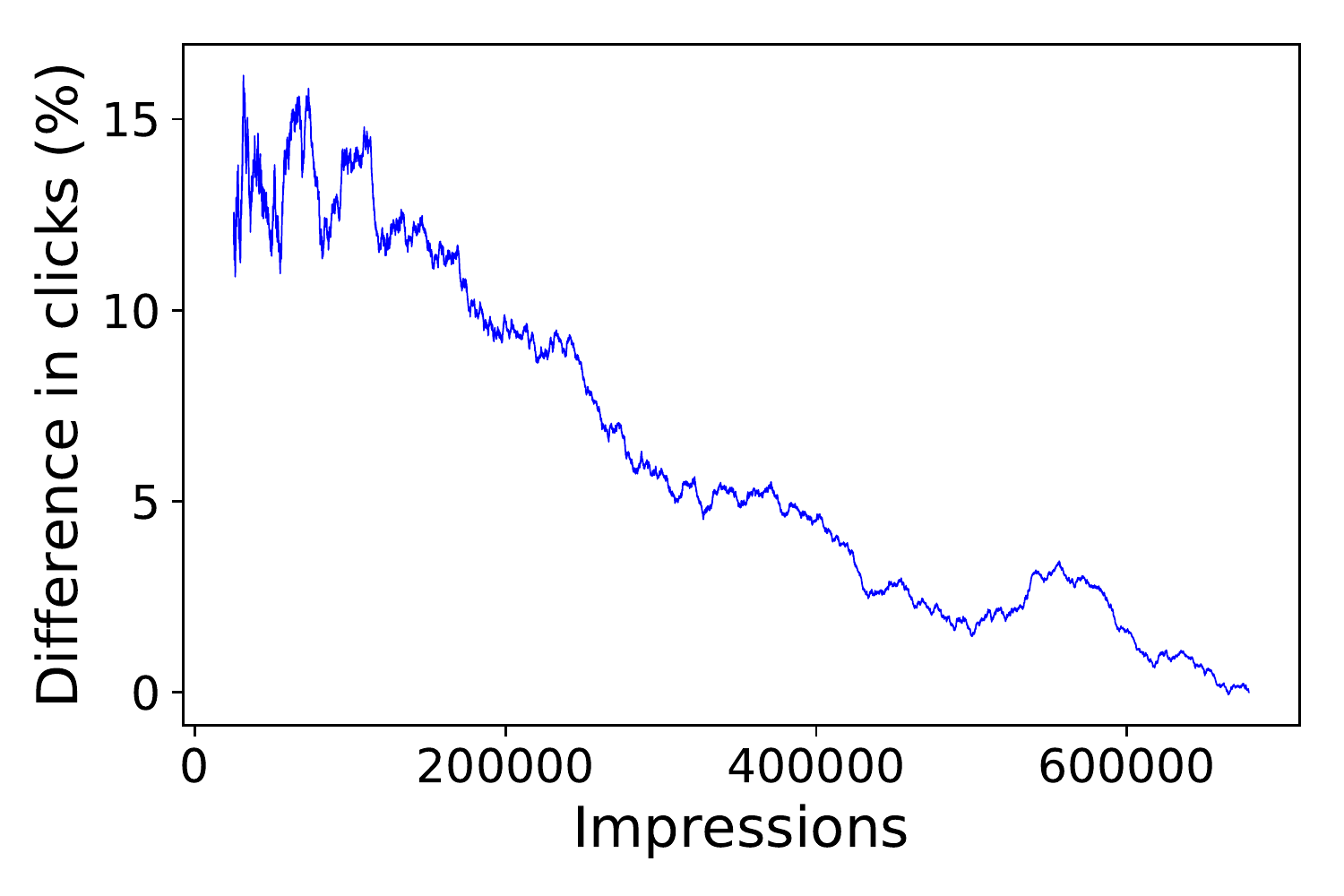}
        \label{fig:FC_difference_in_clicks_percentage}
    \end{subfigure}
    \caption{The difference in cumulative clicks received from FAB-COST and the Laplace bandit. FAB-COST generates over $16.1\%$ more clicks after around $31,000$ impressions.  }\label{fig:Bandit_performance_1}
    
\end{figure}

So that the bandit algorithms could be tested in an offline setting, the following was decided: It was assumed that the action space of available adverts at
the beginning of the simulation $\mathbf{A}^0$ was the entire set of adverts
shown on the day. After iteratively showing adverts and observing if they did
or did not receive a click, they are removed from $\mathbf{A}$ meaning that at
each iteration $\mathbf{A}$ reduces by a row in size. Because there are a finite number of clicks in the
day's worth of data, both bandit algorithms will finish the simulation with an
equal cumulative reward (see Figure \ref{fig:Bandit_performance_1}); in a real-time online environment this will not be the case and the difference in the number
clicks will be expected to increase over time. In the experiments performed it is
assumed that clicks are generated independently with respect to time meaning
that a user would choose to click or not click on an advert irrespective of
when it was shown to them. It is important to reiterate that decrease in the improvement of using FAB-COST over the Laplace bandit over the simulation is due to the finite number of clicks available in the offline dataset. In a true online setting it is likely that FAB-COST will continue to outperform. 

By measuring the difference in clicks received -- as shown in figure
\ref{fig:Bandit_performance_1} -- it is clear that FAB-COST is the better
algorithm, generating over $16\%$ more clicks after around $31,000$
impressions, and given reasonable assumptions about continued new
adverts we might expect such an improvement in performance to persist
and yield significantly improved results over time.

%
%

 \begin{algorithm}
 Set $E$ as the iteration(s) at which you want to make an EP update to the posterior. \\
Initialise the prior distribution $q_0(\bm\theta) = \mathcal{N}(\bm\mu_0, \bm\Sigma_0)$\\
 \For{$ i = 1 \dots T$}
    {
    1. Generate a sample from the approximated posterior:\\ \hspace{1cm}$\tilde{\bm\theta}_i \sim \mathcal{N}(\bm\mu_{i-1},\bm\Sigma_{i-1})$\\
    2. Select an advert: \\
    \hspace{1cm}$a_i = \underset{\mathrm{ j \in \mathcal{A}}}{\text{argmax}} (\mathbf{A}^i \tilde{\bm{\theta}}_i)$\\
    3. Update moments via ADF:\\
    \hspace{1cm}$\bm{\mu}_i  = \bm{\mu}_{i-1} + \bm{\Sigma}_{i-1} \bm{\alpha}_i$\\
    \hspace{1cm}$\bm{\Sigma}_i  = \bm{\Sigma}_{i-1}- \bm{\Sigma}_{i-1}\left(\bm{\alpha}_i \bm{\alpha}_i^\top - 2 \mathbf{B}_i \right)\bm{\Sigma}_{i-1}$ \\
    4. IF i = E , perform an EP approximation as shown in algorithm \ref{alg:EP_calculations} using the last $E$ data points.\\

    }
\caption{FAB-COST}\label{alg:FAB-COST}
\end{algorithm} 

\section{Discussion}

\label{sec:discuss}

In this paper the Fast Approximate Bayesian Cold Start
algorithm FAB-COST has been introduced; a fully Bayesian algorithm which combines both Expectation
Propagation and Assumed Density Filtering to improve on the inference procedure
for the logistic bandit proposed by Chapelle et al.\
\cite{chapelle2011empirical}. Not only would it be beneficial to use
FAB-COST's inference procedure in a bandit setting, but any online learning
scenario for Bayesian logistic regression or with data sets prohibitively large
for EP to be used on.  This is the first time to the authors knowledge that either EP
or ADF have been used to improve bandit performance. FAB-COST addresses two
problems with the classic Laplace approximation: firstly it is an online scheme
which can deal with large cumulative amounts of data and fast throughout;
secondly it achieves better variance accuracy which will result in better
balance between exploration and exploitation and hence improved website
performance, which one would expect to see in a variety of contexts.

%
%

\section*{Acknowledgment}

A big thank you to David Hoyle and Robert Trickey from Auto Trader for their meets and discussions at the beginning of my PhD. Thanks to Edward Pyzer-Knapp from IBM research for his advice on how to take this paper forward. Acknowledgements also go to Simon Barthelm\'e for his emails and discussion on EP.

\ifCLASSOPTIONcaptionsoff
  \newpage
\fi



\bibliographystyle{IEEEtran}
\bibliography{mybib}
%

\end{document}